%% file: main.tex
\documentclass[]{bytedance_seed}

\usepackage[toc,page,header]{appendix}

\input{resources/packages}
\input{resources/math_macro}

\usepackage{amssymb}

\title{VaRL: Value-assisted Reinforcement Learning for Diffusion Post-Training}

\author[1,2]{Hanzhong Guo}
\author[2]{Jie Wu}
\author[3]{Shen Nie}
\author[2]{Zilyu Ye}
\author[1]{Jiahang Cao}
\author[4]{Zhe Zhang}
\author[2]{Qinghao Ye}
\author[2,5]{Jie Liu}
\author[2]{Yu Gao}
\author[2]{Xionghui Wang}
\author[6]{Jean Ponce}
\author[1]{Yizhou Yu}

\affiliation[1]{HKU}
\affiliation[2]{Seed}
\affiliation[3]{RUC}
\affiliation[4]{SUST}
\affiliation[5]{CUHK}
\affiliation[6]{Ecole Normale Sup{\fontencoding{OT1}\fontfamily{cmss}\selectfont\'e}rieure}

\def\contributionlist{{\fontsize{10}{12}\selectfont Jie Wu is the project lead. Yizhou Yu is the corresponding author.}}

\abstract{
Group-wise post-training is widely used to align diffusion models, yet treating rewards purely as scalar weights without differentiating through the sampler makes optimization indirect and sample-inefficient.
Direct reward-gradient methods such as ReFL, DRaFT-LV, DRTune, and LeapAlign improve efficiency by backpropagating differentiable rewards, but rely on truncated or partially detached surrogate paths that deviate from multi-step inference; consequently, increasing reward along the surrogate path does not guarantee higher quality on full inference rollouts.
We propose Value-assisted Reinforcement Learning (VaRL), an on-policy endpoint-distillation framework that learns a differentiable approximation to the current policy's remaining denoising trajectory.
In each round, a frozen snapshot of the current policy generates complete trajectories, from which a student velocity model learns scheduler-consistent endpoint predictions at intermediate states.
The policy is then updated through a single local denoising step composed with the student endpoint map and a differentiable reward model, keeping the backward path short while anchoring supervision to recent full rollouts.
Under local endpoint-accuracy and smoothness conditions, we bound the discrepancy between the student-induced adjoint and the full-rollout reward gradient.
On FLUX.1-dev, VaRL achieves 0.4229 on HPSv2.1 and 78.58 on GenEval while maintaining stable reward growth over more than 10,000 updates.
On Wanx-T2V-14B, our 20-step model achieves aesthetic and motion GSB gains of 44.33\% and 33.33\% over the pretrained model and consistently outperforms a ReFL-style baseline.
Finally, simply averaging the trained teacher and student weights yields competitive 4-step and 8-step generators without extra distillation.
}

\begin{document}
\begin{CJK*}{UTF8}{gbsn}

\maketitle

\begin{figure}[t]
\centering
\includegraphics[
    width=\linewidth,
    trim=14 10 12 10, clip,
    page=1
]{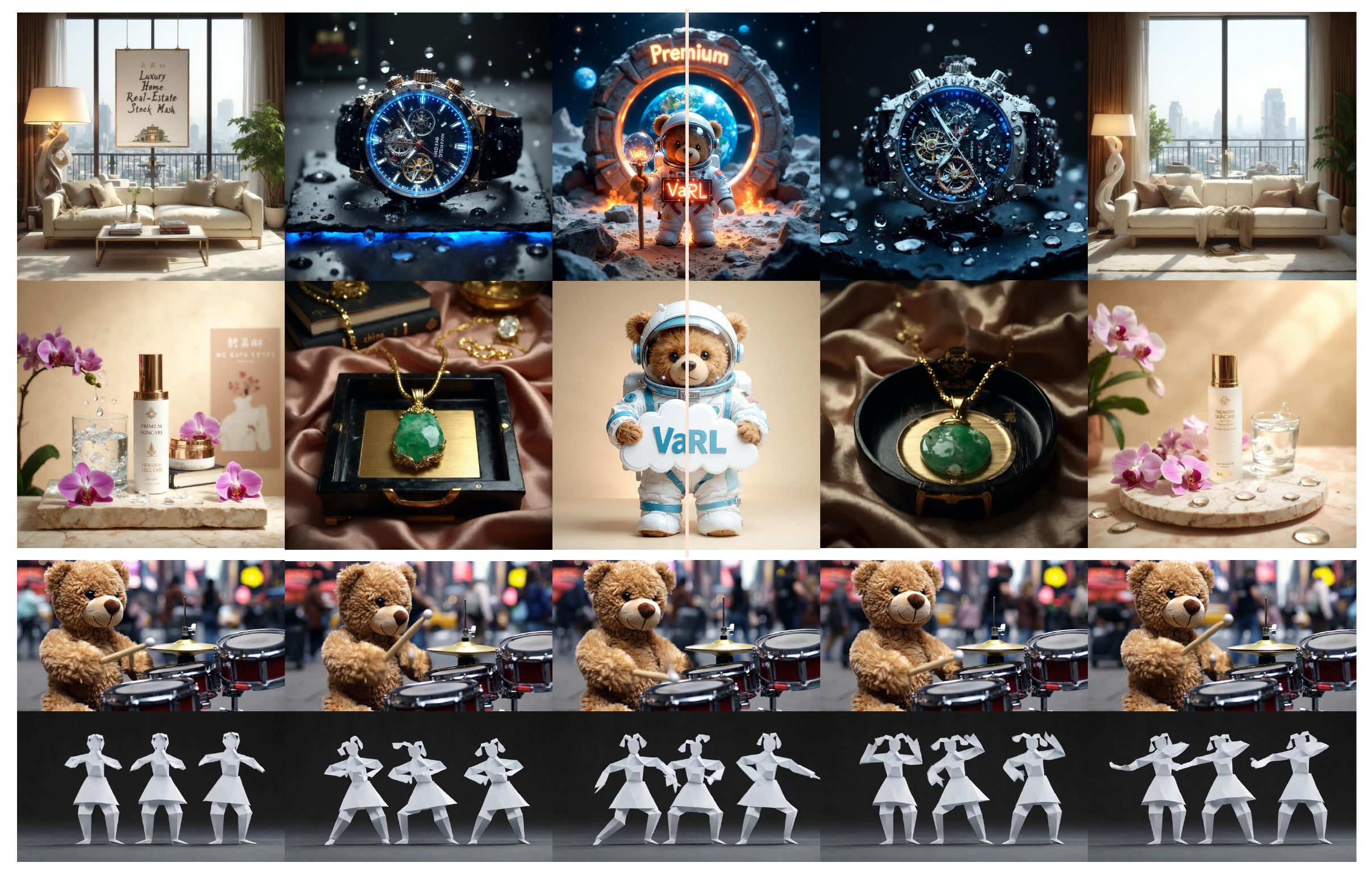}
\vspace{-10pt}
\caption{
\textbf{Images and videos generated by VaRL.}
VaRL is designed to improve the full-step base generator while yielding strong few-step models through direct teacher--student weight averaging.
The top two rows show FLUX.1-dev generations from our post-trained 28-step base policy (left) and its training-free 8-step 50/50 merged counterpart (right); the bottom two rows show video frames from our post-trained 20-step Wanx-T2V-14B policy.
}
\label{fig:demo}
\end{figure}

\input{sections/introduction}
\input{sections/related_work}
\input{sections/method}
\input{sections/experiments}
\input{sections/conclusion}

\bibliographystyle{plainnat}
\bibliography{main}

\clearpage
\appendix
\input{sections/appendix}

\end{CJK*}
\end{document}

%% file: resources/packages.tex
\usepackage{CJKutf8}

\usepackage{xargs}  

\usepackage{todonotes}  

\usepackage{amsmath}
\usepackage{dsfont}

\usepackage{svg}

%% file: resources/math_macro.tex
\newcommand{\refl}{ReFL}

\newcommand{\diffpolicy}{\pi_{\theta}}

\newcommand{\rewardmodel}{R}
\newcommand{\latent}{x}
\newcommand{\decoder}{D}
\newcommand{\studentmap}{G_{\phi}}
\newcommand{\prompt}{c}
\newcommand{\expect}{\mathbb{E}}

%% file: sections/introduction.tex
\section{Introduction}

Diffusion and flow-matching models have become the dominant backbones for high-fidelity image and video generation~\citep{rombach2022latent,saharia2022photorealistic,ramesh2022hierarchical,peebles2023scalable,podell2023sdxl,chen2023pixartalpha,esser2024scaling,flux2024,seedream2025,wu2025qwen,ho2022video,ho2022imagenvideo,singer2022makeavideo,blattmann2023alignyourlatents,yang2024cogvideox,kong2024hunyuanvideo,seedance2025,seedance2026seedance}.
While pretraining captures the underlying data distribution, practical deployment further demands prompt faithfulness, aesthetic quality, and coherent motion aligned with human preference.
Learned reward and preference models, including ImageReward~\citep{xu2023imagereward}, HPS~\citep{wu2023hps,ma2025hpsv3}, PickScore~\citep{kirstain2023pickapic}, UnifiedReward~\citep{unifiedreward2025}, and RewardDance~\citep{rewarddance2025}, make these qualities directly measurable, motivating diffusion reinforcement learning (RL) over multi-step denoising trajectories~\citep{black2023ddpo,fan2024dpok}.

Unlike autoregressive language models, where discrete token sampling blocks pathwise derivatives and favors group-wise estimators such as GRPO~\citep{shao2024grpo}, diffusion samplers evolve through continuous, differentiable latent updates.
Although reward feedback arrives only at the end of a trajectory, a differentiable reward model allows pathwise gradients to propagate directly through the denoising steps (\emph{direct reward gradients}, as opposed to non-differentiating \emph{score-function policy gradients}).
Moreover, diffusion schedulers naturally map intermediate velocity or noise predictions into clean-endpoint estimates $x_0$, motivating methods that decode a predicted endpoint from an intermediate step and backpropagate reward gradients through that shortened path.

Recent visual post-training methods explore both paradigms.
Score-function methods such as DDPO~\citep{black2023ddpo} and DPOK~\citep{fan2024dpok}, together with group-wise approaches including DanceGRPO~\citep{dancegrpo2025}, Flow-GRPO~\citep{liu2025flowgrpo}, DiffusionNFT~\citep{zheng2025diffusionnft}, and Edit-R1~\citep{guo2026editr1}, weight trajectories by scalar rewards without differentiating through the sampler.
Conversely, direct reward-gradient methods such as \refl{}~\citep{xu2023imagereward}, DRaFT-LV~\citep{clark2024direct}, DRTune~\citep{wu2024drtune}, and LeapAlign~\citep{leapalign2026} propagate gradients through predicted endpoints or truncated rollouts, achieving substantially better sample efficiency on large backbones.

The central bottleneck for direct reward-gradient methods is \emph{train--inference inconsistency}.
To keep memory bounded and avoid gradient explosion, \refl{}~\citep{xu2023imagereward} evaluates a one-step endpoint prediction from a late timestep, DRaFT-LV~\citep{clark2024direct} restricts gradients to the final denoising steps, DRTune~\citep{wu2024drtune} detaches intermediate model inputs, and LeapAlign~\citep{leapalign2026} constructs coarse two-step leap trajectories from earlier states.
While computationally practical, none of these surrogate paths matches the multi-step trajectory executed at inference; when the surrogate drifts from the policy's true rollout, optimizing the shortcut reward can degrade full-chain generation over long training horizons.

To bridge this gap, we introduce \textbf{Value-assisted Reinforcement Learning (VaRL)}, an on-policy endpoint-distillation framework aimed at raising the post-training ceiling of full-step diffusion generators (Fig.~\ref{fig:demo}).
In each round, a frozen copy of the current policy rolls out complete trajectories into a latest-rollout buffer, supervising an online student velocity model whose scheduler-induced endpoint map approximates the terminal latent reached by the full chain.
The policy then takes one differentiable denoising step followed by the student endpoint map and reward model, keeping gradient depth short while anchoring updates to full rollouts.
While our core objective is improving the full-step base model rather than training a standalone few-step student, the student absorbs multi-step teacher dynamics along visited states, enabling 50/50 teacher--student weight averaging to yield strong 4- and 8-step generators without extra distillation.

Our main contributions are threefold:
(1) We propose VaRL, coupling on-policy endpoint distillation from complete teacher rollouts with a short, staged teacher--student reward-gradient update.
(2) Under local endpoint-accuracy and smoothness conditions, we show that the student-induced adjoint approximates the continuous full-rollout reward gradient within $O(\epsilon_0+\sqrt{\epsilon_{\mathrm{loc}}})$.
(3) Across FLUX.1-dev and Wanx-T2V-14B, VaRL delivers stable reward improvements over long training horizons, while a 50/50 merge of the final teacher and student weights yields competitive 4- and 8-step models without extra distillation.

%% file: sections/related_work.tex
\section{Related Work}

\paragraph{Post-training in language models.}
Preference alignment over discrete token sequences typically fits a reward model followed by PPO~\citep{ouyang2022training,schulman2017ppo}, optimizes directly on preference pairs via DPO~\citep{rafailov2023direct}, or replaces learned critics with group-relative advantages in GRPO~\citep{shao2024grpo}.
In parallel, knowledge, policy, and on-policy distillation transfer stronger teacher behaviors into compact or stable students~\citep{hinton2015distilling,rusu2015policy,rethinkingopd2026,opdsurvey2026}.
Because discrete token sampling blocks pathwise derivatives, these methods cannot differentiate through generated outputs, whereas continuous diffusion samplers allow differentiable reward models to provide pathwise gradients through synthesized images or videos.

\paragraph{Post-training in diffusion and visual generation.}
Diffusion RL casts multi-step denoising as a sequential decision process maximizing a reward on the final decoded output.
Early policy-gradient methods such as DDPO and DPOK estimate score-function gradients over complete denoising trajectories~\citep{black2023ddpo,fan2024dpok}, while DPPO studies policy-gradient fine-tuning for continuous-control diffusion policies and highlights how diffusion parameterizations support structured exploration~\citep{ren2025dppo}.
Preference-based objectives adapt DPO to diffusion likelihoods or trajectories to reduce reliance on explicit reward models~\citep{wallace2024diffusiondpo,yang2024d3po}, and recent group-wise methods bring GRPO-style training to visual generation: DanceGRPO~\citep{dancegrpo2025} applies group-relative updates to image and video backbones, Flow-GRPO~\citep{liu2025flowgrpo} converts deterministic flow-matching samplers into stochastic SDEs for online exploration, and Edit-R1~\citep{guo2026editr1} introduces verifier-based reward modeling for image editing.
Across these paradigms, alignment quality relies on capable preference and multimodal reward models, including ImageReward~\citep{xu2023imagereward}, HPSv2~\citep{wu2023hps}, HPSv3~\citep{ma2025hpsv3}, PickScore~\citep{kirstain2023pickapic}, UnifiedReward~\citep{unifiedreward2025}, and RewardDance~\citep{rewarddance2025}.
Unlike discrete token samplers, diffusion and flow models are differentiable from intermediate latents to endpoint predictions and decoded pixels, enabling direct reward-gradient methods such as \refl{}~\citep{xu2023imagereward}, AlignProp~\citep{prabhudesai2023alignprop}, DRaFT/DRaFT-LV~\citep{clark2024direct}, DRTune~\citep{wu2024drtune}, and LeapAlign~\citep{leapalign2026}.
However, because these methods rely on one-step endpoint estimates, truncated backpropagation windows, or detached inputs to stay within memory limits, their training gradients can diverge from the full multi-step inference chain.
Meanwhile, diffusion distillation typically compresses multi-step teachers into few-step generators~\citep{salimans2022progressive,song2023consistency,luo2023latent,sauer2024adversarial,yin2024improved}, and recent works such as PSO~\citep{miao2024pso}, DMDR~\citep{jiang2025dmdr}, Hyper-SD~\citep{ren2024hypersd}, and Diff-Instruct++~\citep{luo2024diffinstructpp} apply RL or preference optimization directly to align standalone few-step distilled models.
Our focus is complementary: VaRL is primarily designed to improve the full-step base model by using an on-policy student endpoint map as a rollout-consistent differentiable surrogate during policy updates, while simple parameter merging of the trained teacher and student still yields competitive 4- and 8-step models without dedicated step-distillation training.

%% file: sections/method.tex
\section{Method}
\label{sec:method}

\begin{figure}[t]
\centering
\includegraphics[
    width=\linewidth,
    page=1
]{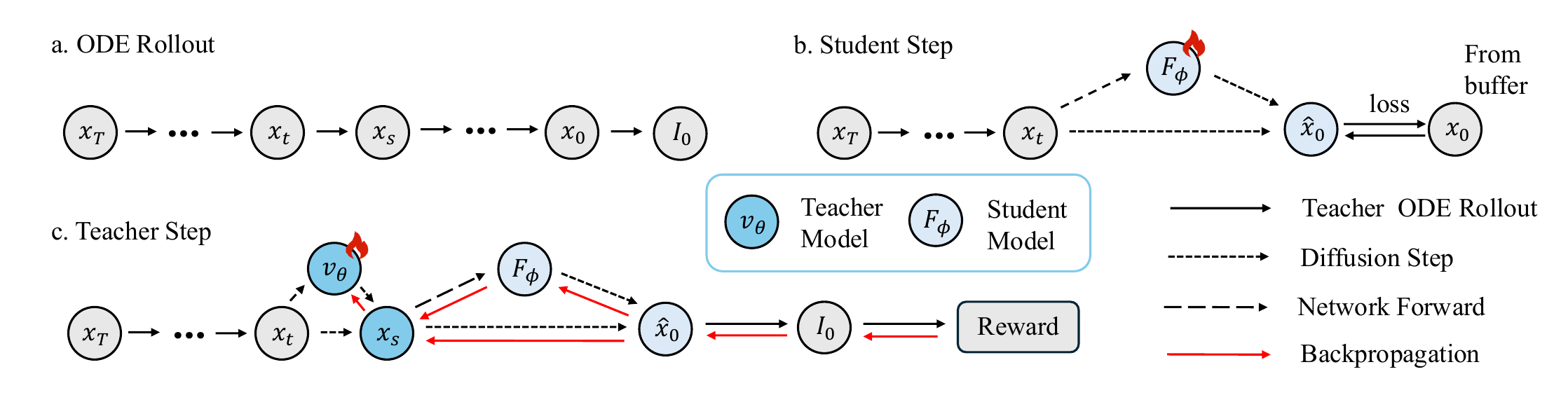}
\caption{
\textbf{Training pipeline of VaRL.}
\textbf{a}, A stop-gradient teacher rollout generates a complete ODE trajectory and caches intermediate states along with the terminal latent $x_0$ in a buffer holding the latest rollouts.
\textbf{b}, Drawing intermediate states and endpoint targets from this latest-rollout buffer, the student velocity model $F_{\phi}$ predicts a velocity that the scheduler converts into an endpoint estimate supervised by endpoint and velocity losses.
\textbf{c}, During policy optimization, the trainable policy $v_{\theta}$ takes one local denoising step from $\latent_t$ to $\latent_s$ followed by the student endpoint map, restricting backpropagation to this short differentiable path.
}
\label{fig:teaser}
\end{figure}

\subsection{Diffusion RL Setup}

Let $\prompt$ denote the conditioning text prompt.
Over a decreasing time schedule from noise ($t=T$) to clean data ($t=0$), a diffusion or flow-matching model defines a denoising policy $\diffpolicy$ generating a trajectory
$\tau = \{\latent_T, \ldots, \latent_0\}$ initialized at $\latent_T\sim\mathcal{N}(0,I)$.
The terminal latent $\latent_0$ is decoded into an image or video $I_0=\decoder(\latent_0)$ and scored by a differentiable reward model $\rewardmodel(I_0,\prompt)$:
\begin{equation}
    \max_{\theta} \; \expect_{\tau \sim \diffpolicy(\cdot|\prompt)}[\rewardmodel(\decoder(\latent_0),\prompt)].
\end{equation}
Because the reward is evaluated only at $\latent_0$, backpropagating through the entire multi-step chain is memory-prohibitive, whereas optimizing a naive truncated shortcut introduces a mismatch with multi-step inference.

\subsection{On-Policy Distillation for Diffusion}

Rather than fitting a scalar critic on noisy states, VaRL distills the remaining trajectory into an endpoint predictor.
At each training step, a frozen snapshot $\pi_{\bar{\theta}}$ of the current policy rolls out a complete stop-gradient trajectory over $t\in[0,T]$.
Let $\Phi_{\bar{\theta}}^{t\rightarrow 0}$ denote the teacher rollout map from an intermediate state $\latent_t$ to its terminal denoised latent:
\begin{equation}
    x^{\mathrm{ref}}_{0}
    =
    \Phi_{\bar{\theta}}^{t\rightarrow 0}(\latent_t,\prompt).
\end{equation}
Since differentiating through $\Phi_{\bar{\theta}}^{t\rightarrow 0}$ requires unrolling the remaining chain, we train an online student velocity model $F_{\phi}(\latent_t,t,\prompt)$ along states visited by the teacher rollout and convert its output into an endpoint prediction via the scheduler map $\mathcal{S}_{0,t}$:
\begin{equation}
    \hat{v}_{\phi}
    =
    F_{\phi}(\latent_t,t,\prompt),
    \qquad
    \hat{x}^{\phi}_0
    =
    \studentmap(\latent_t,t,\prompt)
    =
    \mathcal{S}_{0,t}\!\left(\latent_t,\hat{v}_{\phi}\right).
\end{equation}
Under the linear flow interpolation $\latent_t=(1-t)\latent_0+t\epsilon$ with $\epsilon\sim\mathcal{N}(0,I)$ and velocity $v=\epsilon-\latent_0$, this endpoint conversion takes the explicit form
\begin{equation}
    \mathcal{S}_{0,t}(\latent_t,v)=\latent_t-tv.
\end{equation}
Given the rollout endpoint $x_0^{\mathrm{ref}}=\Phi_{\bar{\theta}}^{t\rightarrow 0}(\latent_t,\prompt)$, the corresponding chord velocity target over $[0,t]$ ($t>0$) is
\begin{equation}
    v_t^{\mathrm{ref}}
    =
    \frac{\latent_t-x_0^{\mathrm{ref}}}{t},
\end{equation}
representing the average velocity mapped back to $x_0^{\mathrm{ref}}$ by $\mathcal{S}_{0,t}$ rather than the teacher's instantaneous velocity~\citep{liu2022flow,geng2025meanflow}.
Wrapping rollout targets with the stop-gradient operator $\mathrm{sg}(\cdot)$ so the multi-step teacher trajectory stays outside the computation graph, we supervise $F_{\phi}$ with both endpoint and velocity regression:
\begin{align}
    \mathcal{L}_{x_0}
    &=
    \left\|
    \hat{x}^{\phi}_0
    -
    \mathrm{sg}(x_0^{\mathrm{ref}})
    \right\|_2^2,\\
    \mathcal{L}_{v}
    &=
    \left\|
    F_{\phi}(\latent_t,t,\prompt)
    -
    \mathrm{sg}\!\left(v_t^{\mathrm{ref}}\right)
    \right\|_2^2.
\end{align}
For adjacent states $(\latent_t,\latent_s)$ from the same teacher rollout, we additionally define
\begin{align}
    \mathcal{L}_{\mathrm{LCM}}
    &=
    \left\|
    \studentmap(\latent_t,t,\prompt)
    -
    \mathrm{sg}\!\left(\studentmap(\latent_s,s,\prompt)\right)
    \right\|_2^2,\\
    \mathcal{L}_{\mathrm{LPIPS}}
    &=
    \operatorname{LPIPS}\!\left(
    \decoder(\hat{x}^{\phi}_0),
    \decoder(\mathrm{sg}(x_0^{\mathrm{ref}}))
    \right).
\end{align}
The student objective is
\begin{equation}
    \mathcal{L}_{\mathrm{OPD}}(\phi)
    =
    \expect_{\tau\sim\pi_{\bar{\theta}}}
    \bigl[
    \lambda_{x_0}\mathcal{L}_{x_0}
    +
    \lambda_v\mathcal{L}_{v}
    +
    \lambda_{\mathrm{LCM}}\mathcal{L}_{\mathrm{LCM}}
    +
    \lambda_{\mathrm{LPIPS}}\mathcal{L}_{\mathrm{LPIPS}}
    \bigr].
\end{equation}
Although $\mathcal{L}_{x_0}$ and $\mathcal{L}_v$ share the target $x_0^{\mathrm{ref}}$ and differ only by a time-dependent scaling under linear flow, retaining both allows flexible timestep weighting; backbone-specific choices for $\lambda_{\mathrm{LCM}}$ and $\lambda_{\mathrm{LPIPS}}$ are detailed in Sec.~\ref{sec:experiments}.
In implementation, intermediate states and terminal targets $x_0^{\mathrm{ref}}$ from the latest teacher rollouts are cached in a temporary on-policy buffer (labeled ``From buffer'' in Fig.~\ref{fig:teaser}b) for student updates and refreshed on subsequent rollouts, avoiding stale trajectories from earlier policy iterations.
Proposition~1 formalizes why this endpoint supervision aids policy optimization: when $\studentmap$ is locally accurate and smooth around visited rollout states, the resulting student-induced adjoint closely tracks the full-rollout reward gradient.

\begin{tcolorbox}[
  title=\textbf{Proposition 1: Conditional Adjoint Alignment from OPD},
  width=\linewidth,
  breakable,
  colback=gray!4,
  colframe=seedblue,
  boxrule=0.7pt,
  arc=2pt
]
Consider the continuous-time sampler $d\latent_t/dt=v_{\theta}(\latent_t,t,\prompt)$ with decoded terminal reward $\rewardmodel(\decoder(\latent_0),\prompt)$.
The full rollout gradient can be written as the adjoint integral
\begin{equation}
    \nabla_{\theta}J
    =
    \int_T^0
    \left(\partial_{\theta}v_{\theta}(\latent_t,t,\prompt)\right)^{\top}
    a(t)\,dt ,
\end{equation}
where $a(t)$ is the reward adjoint propagated from the decoded terminal output back to time $t$.
VaRL uses the composed student endpoint map to approximate this adjoint:
\begin{equation}
    \hat{a}_{\phi}(t)
    =
    \left(\nabla_{\latent}\studentmap(\latent_t,t,\prompt)\right)^{\top}
    \nabla_{\latent_0}
    \left[
    \rewardmodel(\decoder(\latent_0),\prompt)
    \right]_{\latent_0=\studentmap(\latent_t,t,\prompt)}.
\end{equation}
When $\bar{\theta}=\theta$, if the composed student endpoint map matches the continuous solution map from time $t$ to the endpoint latent at the sampled state and over a small on-policy neighborhood, then
\begin{equation}
    \left\|
    \hat{a}_{\phi}(t)-a(t)
    \right\|
    =
    O(\epsilon_0+\sqrt{\epsilon_{\mathrm{loc}}}),
\end{equation}
where $\epsilon_0$ measures endpoint error at the sampled state, including the solver gap between the discrete and continuous solution maps, and $\epsilon_{\mathrm{loc}}$ measures neighborhood endpoint error. Local smoothness constants are absorbed into the $O(\cdot)$ notation.
Consequently, if $g_{\mathrm{varl}}(t)=-(\partial_{\theta}v_{\theta})^{\top}\hat{a}_{\phi}(t)$ and $t$ is sampled uniformly from $[0,T]$, then
\begin{equation}
    \left\|
    T\,\expect_t[g_{\mathrm{varl}}(t)]
    -
    \nabla_{\theta}J
    \right\|
    \leq
    C\left(\epsilon_0+\sqrt{\epsilon_{\mathrm{loc}}}\right).
\end{equation}
Here $C$ absorbs local smoothness constants.
Sec.~\ref{sec:policy_update} and Appendix~\ref{app:gradient_alignment_proof} detail how the discrete one-step policy transition realizes this integrand on non-uniform schedules, together with the full proof and regularity assumptions.
\end{tcolorbox}

\subsection{Policy Update Through the Student Endpoint}
\label{sec:policy_update}

Because $\studentmap$ is parameterized solely by $\phi$, updating the policy parameters $\theta$ requires routing a differentiable transition from $\theta$ into the student endpoint map.
On the discrete sampling schedule $T=t_N > t_{N-1} > \cdots > t_0=0$, given a sampled rollout state $\latent_t$ at $t=t_k$ and the next schedule time $s=t_{k-1}$ with step size $\Delta t = t-s > 0$, the trainable policy executes one local denoising step before querying the student endpoint predictor and decoder:
\begin{equation}
    \hat{v}_{\theta}=v_{\theta}(\latent_t,t,\prompt),\quad
    \latent_{s}^{\theta}=\mathcal{S}_{s,t}(\latent_t,\hat{v}_{\theta}),\quad
    \hat{v}_{\phi}=F_{\phi}(\latent_{s}^{\theta},s,\prompt),\quad
    \hat{x}_0=\mathcal{S}_{0,s}(\latent_{s}^{\theta},\hat{v}_{\phi}),\quad
    \hat{I}_0=\decoder(\hat{x}_0),
\end{equation}
where $\mathcal{S}_{s,t}(\latent_t,\hat{v}_{\theta})=\latent_t-\Delta t\,\hat{v}_{\theta}$ is the one-step Euler flow update and $\mathcal{S}_{0,s}$ is the clean-endpoint projection at time $s$.
Applying the chain rule through $\partial \mathcal{S}_{s,t}/\partial \hat{v}_{\theta}=-\Delta t\,I$ gives the local reward gradient
\begin{equation}
    \nabla_{\theta}\rewardmodel(\hat{I}_0,\prompt)
    =
    -\Delta t\left(\partial_{\theta}v_{\theta}(\latent_t,t,\prompt)\right)^{\top}\hat{a}_{\phi}^{\theta}(s),
\end{equation}
where $\hat{a}_{\phi}^{\theta}(s)=(\nabla_{\latent}\studentmap(\latent_{s}^{\theta},s,\prompt))^{\top}g_{x_0}$ with $g_{x_0}=\nabla_{\latent_0}[\rewardmodel(\decoder(\latent_0),\prompt)]_{\latent_0=\hat{x}_0}$ evaluates the student adjoint at the transitioned state $\latent_{s}^{\theta}$.
Here, the scheduler Jacobian factor $\Delta t$ naturally supplies the non-uniform (or flow-shifted) quadrature weight for the adjoint integral in Proposition~1, while at $\bar{\theta}=\theta$ the post-step state $\latent_{s}^{\theta}=\latent_s$ lies directly on the rollout trajectory and shifts the adjoint evaluation from $t$ to $s$ only by an $O(\Delta t)$ discretization step (Appendix~\ref{app:gradient_alignment_proof}).

Following \refl{}, we clip the scalar reward via $\widetilde{\rewardmodel}=\operatorname{clip}(\rewardmodel,r_{\min},r_{\max})$ for stability and optimize $\theta$ with
\begin{equation}
    \mathcal{L}_{\mathrm{VaRL}}(\theta)
    =
    -\widetilde{\rewardmodel}(\hat{I}_0,\prompt)
    +
    \beta_{\mathrm{KL}}\mathcal{L}_{\mathrm{KL}}(\theta,\theta_{\mathrm{ref}}),
\end{equation}
where $\mathcal{L}_{\mathrm{KL}}(\theta,\theta_{\mathrm{ref}})$ regularizes policy drift against the frozen reference model $\theta_{\mathrm{ref}}$ with backbone-dependent weight $\beta_{\mathrm{KL}}$ (differentiated separately from the reward path).
Both backbones retain their default deterministic schedulers and generate teacher rollouts under classifier-free guidance (one conditional and one unconditional pass per step).
Rather than sampling $t$ uniformly over $[0,T]$, policy updates draw timesteps from the high-noise subset $\mathcal{H}$ of the schedule.
High-noise steps govern global composition, counting, spatial layout, and motion structure, whereas near $t=0$ structure is largely fixed and one-step endpoint estimates are already accurate; one-step shortcuts like \refl{} break down at high noise and are restricted to late steps.
With $\studentmap$ distilling the remaining rollout, each update at $k\in\mathcal{H}$ approximates that step's exact contribution $g_k$ to the full rollout gradient $\nabla_{\theta}J_h=\sum_{k=1}^{N}g_k$, effectively applying full-rollout reward gradients selectively to the high-noise regime (and recovering Proposition~1 across $[0,T]$ under uniform sampling).

To avoid holding the computation graphs of the reward model, VAE decoder, student velocity model, and trainable policy in GPU memory simultaneously, we execute backpropagation along this short path in stages.
Specifically, we first run the local teacher step, the student endpoint projection, the latent decoder, and the reward model without retaining intermediate activations, caching the terminal output gradient
\begin{equation}
    g_I=\nabla_{\hat{I}_0}\widetilde{\rewardmodel}(\hat{I}_0,\prompt).
\end{equation}
Treating $g_I$ as a cached upstream cotangent, we then recompute and differentiate each lightweight stage sequentially in reverse order: backpropagating $g_I$ through $\decoder$ onto $\hat{x}_0$, replaying $\studentmap$ onto $\latent_{s}^{\theta}$, differentiating $\mathcal{S}_{s,t}$ onto $\hat{v}_{\theta}$, and finally running one differentiated forward pass of $v_{\theta}$ to update $\theta$.
Because each stage uses identical deterministic forward inputs, this staged replay recovers the exact local reward gradient while keeping peak activation memory close to single-module differentiation (with frame subsampling additionally applied in our video experiments; see Sec.~\ref{sec:experiments}).

\subsection{Train-Inference Consistency}

Continually refitting $\studentmap$ on current stop-gradient rollouts keeps the surrogate endpoint aligned with what the multi-step policy actually generates at inference.
Unlike static one-step shortcuts, which can be exploited along directions that fail to survive full-chain denoising, routing gradients through an on-policy student constrains local policy updates toward endpoints reachable by the full sampler.
Empirically, both the short-path training reward and the held-out full-sampler evaluation reward improve steadily for over 10,000 updates while preserving visual coherence.

\subsection{Student-Reward Value Proxy}

Although VaRL does not train an explicit scalar critic, composing the student endpoint map with the decoder and reward model naturally induces a state-value proxy~\citep{sutton2018reinforcement}:
\begin{equation}
    V_{\phi}(\latent_t,t,\prompt)
    =
    \rewardmodel(\decoder(\studentmap(\latent_t,t,\prompt)),\prompt).
\end{equation}
Whereas actor--critic methods such as PPO~\citep{schulman2017ppo} attach a scalar regression head to predict terminal return directly from noisy latents, VaRL first predicts a clean endpoint latent and evaluates it through $\rewardmodel\circ\decoder$.
We assess return-prediction quality via explained variance,
\begin{equation}
    \mathrm{EV}(V) = 1 - \frac{\mathrm{Var}(r - V)}{\mathrm{Var}(r)} ,
\end{equation}
where $r$ denotes the terminal rollout reward.
In our diagnostic experiment, a PPO-style scalar value head plateaus near $0.4$ explained variance, whereas the endpoint proxy starts near $0.5$ at initialization and improves further with distillation, validating endpoint prediction as an intermediate representation.

\subsection{Few-Step Byproduct via Model Merge}

VaRL is designed primarily to strengthen the full-step base policy $\diffpolicy$, using $F_{\phi}$ as an auxiliary endpoint predictor rather than training a standalone few-step generator from pure noise; nevertheless, $F_{\phi}$ encodes a compressed representation of the teacher's multi-step trajectory.
We obtain few-step generators without any post-hoc distillation by uniformly averaging corresponding parameters of the final teacher and student checkpoints across all layers:
\begin{equation}
    W_{\mathrm{merge}}
    =
    0.5\,W_{\mathrm{teacher}}
    +
    0.5\,W_{\mathrm{student}}.
\end{equation}
Both our 4-step and 8-step evaluations share this single merged checkpoint, changing only the inference step count while setting the guidance scale near $1$ since classifier-free guidance is already distilled into the student velocity.

%% file: sections/experiments.tex
\section{Experiments}
\label{sec:experiments}

\subsection{Experimental Setup}

\paragraph{Pretrained models.}
We evaluate VaRL across both image and video reward-optimization tasks, adopting FLUX.1-dev~\citep{flux2024} as the base text-to-image generator and Wanx-T2V-14B~\citep{wan2025wan} as the base text-to-video generator.
Within each modality, our method and all reproduced baselines are initialized from the exact same pretrained checkpoint.

\paragraph{Reward models.}
For public image preference alignment, we optimize HPSv2.1 on 50,000 prompts sampled from the HPDv2 dataset and evaluate the resulting model on both in-domain HPSv2.1 and held-out HPSv3~\citep{wu2023hps,ma2025hpsv3}.
For public compositional alignment, following the protocol of LeapAlign~\citep{leapalign2026}, we perform a separate HPSv2.1 training run on 50,000 prompts generated with the official GenEval scripts.
Beyond HPSv2.1, to test scalability with large multimodal verifiers, we optimize FLUX.1-dev against a differentiable 26B vision-language reward model, SeedVL from Seed-1.5-VL~\citep{seed15vl2025} paired with expert-written rubrics~\citep{guo2026editr1}, on our internal text-alignment benchmark and compare directly with gradient-based baselines.
For video generation, we train an InternVL-based video reward model~\citep{chen2024internvl} on internally annotated multi-dimensional preference data; following VideoAlign~\citep{liu2025videoreward}, it scores five axes (lighting, texture, motion stability, motion vividness, and prompt following), which are averaged with equal weights into the scalar training reward.

\paragraph{Baselines and implementation.}
For image generation, we compare against both group-wise policy-gradient methods, including DanceGRPO~\citep{dancegrpo2025} and MixGRPO~\citep{li2025mixgrpo}, and direct reward-gradient methods, including \refl{}~\citep{xu2023imagereward}, DRaFT-LV~\citep{clark2024direct}, DRTune~\citep{wu2024drtune}, and LeapAlign~\citep{leapalign2026}.
On the internal VLM text-alignment benchmark, we compare the pretrained FLUX.1-dev model, a \refl{}-style one-step gradient baseline, LeapAlign, and VaRL under identical 26B SeedVL supervision; on Wanx-T2V-14B, we conduct pairwise evaluations against both the pretrained backbone and a \refl{}-style baseline.
VaRL alternates 1:1 between full-rollout student distillation and local policy updates; while rolling out full trajectories and updating an online student makes each iteration slower than lightweight shortcuts such as \refl{}, anchoring updates to full rollouts eliminates shortcut drift and raises the alignment ceiling beyond both truncated-gradient and group-wise RL methods.
FLUX.1-dev enables both consistency and perceptual student losses ($\lambda_{\mathrm{LCM}}=0.5$, $\lambda_{\mathrm{LPIPS}}=1.0$) alongside $\beta_{\mathrm{KL}}=0.004$, whereas Wanx-T2V-14B sets these auxiliary and reference weights to zero ($\lambda_{\mathrm{LCM}}=\lambda_{\mathrm{LPIPS}}=\beta_{\mathrm{KL}}=0$)~\citep{song2023consistency,luo2023latent}.

\paragraph{Evaluation metrics.}
We report HPSv2.1 and HPSv3 for general image preference, and GenEval~\citep{ghosh2024geneval} overall accuracy together with all six task categories (single object, two objects, counting, colors, spatial position, and attribute binding) for compositional alignment.
On the internal text-alignment set of 150 diverse prompts, we sample four images per prompt and score each image with SeedVL against prompt-specific binary rubric items~\citep{seed15vl2025,guo2026editr1}, reporting the mean score across all prompts and rubric points.
For video generation, we report human pairwise Good--Same--Bad (GSB) preference ratios $(G-B)/(G+S+B)$ on prompts from VBench~\citep{huang2024vbench}, where $G$, $S$, and $B$ count wins, ties, and losses relative to the reference model.

\subsection{Reward and Value Analysis}

\begin{figure}[!htbp]
\centering
\includegraphics[width=\linewidth]{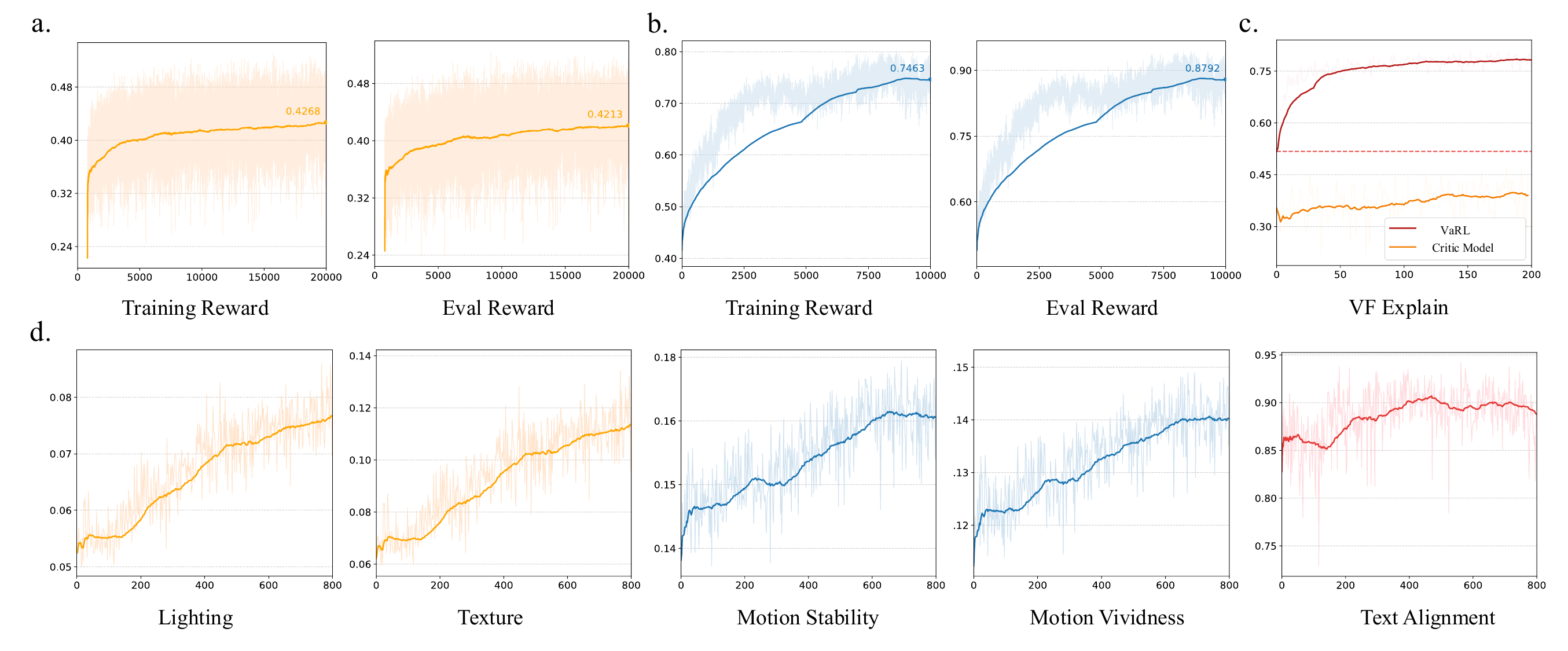}
\caption{
\textbf{Reward dynamics and value-proxy analysis across backbones and reward models.}
\textbf{a,b}, FLUX.1-dev training reward (evaluated along the local teacher--student update path on training prompts) and held-out evaluation reward (evaluated with full teacher rollouts on unseen prompts) under HPSv2.1 (\textbf{a}) and the 26B VLM reward (\textbf{b}).
\textbf{c}, Explained variance of terminal-return prediction from intermediate noisy states, comparing a PPO-style scalar value head against the VaRL student-endpoint value proxy.
\textbf{d}, Per-dimension Wanx-T2V-14B training curves across the five InternVL video reward axes (translucent: raw values; solid: smoothed trends).
}
\label{fig:reward_value_analysis}
\end{figure}

Fig.~\ref{fig:reward_value_analysis} examines reward optimization trajectories and intermediate return-prediction accuracy, measuring training reward along the local update path on training prompts and evaluation reward with complete teacher rollouts on held-out prompts.
On FLUX.1-dev, both training and full-rollout evaluation rewards rise steadily over extended training under HPSv2.1 and the 26B VLM reward (Figs.~\ref{fig:reward_value_analysis}a,b), confirming that surrogate gains transfer reliably to full inference.
In the return-prediction diagnostic (Fig.~\ref{fig:reward_value_analysis}c, defined in Sec.~\ref{sec:method}), a PPO-like baseline that attaches trainable scalar value heads to regress terminal rewards from noisy intermediate latents reaches an explained variance of roughly $0.4$, whereas evaluating student-predicted endpoints through the reward model starts higher and yields substantially stronger return-prediction accuracy.
A similar trend appears at a finer granularity on Wanx-T2V-14B (Fig.~\ref{fig:reward_value_analysis}d), where all five InternVL reward dimensions (lighting, texture, motion stability, motion vividness, and text alignment) improve jointly across the run rather than trading off against one another.

\subsection{HPS and GenEval Results}

\begin{table}[!htbp]
\centering
\caption{
Quantitative comparison of post-training methods on FLUX.1-dev (baseline results cited from prior work).
HPSv2.1 (in-domain) and HPSv3 (held-out) are evaluated from the HPSv2.1-trained checkpoint, while GenEval (0--100 scale) comes from a separate GenEval-prompt run.
VaRL (model merged) uses a single 50/50 teacher--student checkpoint evaluated at 4 steps for HPSv2.1/v3 and 8 steps for GenEval.
Best results are in \textbf{bold} and second-best are \underline{underlined}.
}
\label{tab:hps_geneval_opdo}
\renewcommand{\arraystretch}{1.08}
\resizebox{\textwidth}{!}{
\begin{tabular}{lcc|ccccccc}
\toprule
\multicolumn{1}{c}{}
& \multicolumn{1}{c}{\textbf{In-Domain}}
& \multicolumn{1}{c|}{\textbf{Held-out}}
& \multicolumn{7}{c}{\textbf{GenEval Benchmark}} \\
\cmidrule(lr){2-2}\cmidrule(lr){3-3}\cmidrule(lr){4-10}
\textbf{Method} & \textbf{HPSv2.1$\uparrow$} & \textbf{HPSv3$\uparrow$} & \textbf{Overall$\uparrow$} & \textbf{Single Obj.$\uparrow$} & \textbf{Two Obj.$\uparrow$} & \textbf{Count$\uparrow$} & \textbf{Color$\uparrow$} & \textbf{Pos$\uparrow$} & \textbf{AttrB$\uparrow$} \\
\midrule
\multicolumn{10}{c}{\itshape Pretrained Model} \\
\midrule
FLUX.1-dev & 0.3078 & 13.5020 & 65.35 & 99.38 & 86.62 & 66.88 & 74.47 & 19.50 & 45.25 \\
\midrule
\multicolumn{10}{c}{\itshape Policy-Gradient-Based Methods} \\
\midrule
DanceGRPO & 0.3451 & 14.8336 & 67.75 & 99.38 & 90.15 & 69.38 & 76.33 & 22.25 & 49.00 \\
MixGRPO & 0.3692 & 14.7530 & 72.32 & \underline{99.69} & \underline{93.69} & \underline{80.00} & 80.05 & 24.25 & 56.25 \\
\midrule
\multicolumn{10}{c}{\itshape Direct-Gradient Methods} \\
\midrule
\refl{} & 0.3852 & 15.5127 & 70.11 & 99.38 & 92.68 & 69.06 & 75.80 & 26.75 & 57.00 \\
DRaFT-LV & 0.3859 & 15.3699 & 70.24 & \underline{99.69} & 92.42 & 74.06 & 75.53 & 24.00 & 55.75 \\
DRTune & 0.3882 & 15.5606 & 71.01 & 99.38 & \underline{93.69} & 73.12 & 76.86 & 27.50 & 55.50 \\
LeapAlign & 0.4092 & 15.7678 & 74.20 & 99.38 & \textbf{96.46} & 72.50 & 80.59 & 30.25 & 66.00 \\
VaRL & \textbf{0.4229} & \textbf{15.8629} & \underline{78.58} & \textbf{100.00} & 93.18 & \textbf{84.38} & \textbf{86.44} & \underline{36.00} & \underline{71.50} \\
VaRL (model merged) & \underline{0.4212} & \underline{15.8522} & \textbf{78.94} & \underline{99.69} & 92.42 & 78.12 & \underline{86.17} & \textbf{39.50} & \textbf{77.75} \\
\bottomrule
\end{tabular}
}
\end{table}

Tab.~\ref{tab:hps_geneval_opdo} summarizes public preference and compositional alignment benchmarks on FLUX.1-dev, with baseline scores cited from prior reports under their standard splits.
VaRL outperforms both group-wise policy-gradient and prior direct-gradient baselines on in-domain HPSv2.1 as well as held-out HPSv3, while delivering substantial GenEval gains on counting, spatial position, and attribute binding.
Moreover, the training-free 50/50 teacher--student merge, evaluated at 4 steps on HPS and 8 steps on GenEval, closely matches the 28-step teacher on preference metrics and achieves the highest overall GenEval score (78.94).

\subsection{FLUX.1-dev Image-Text Alignment Results}

\begin{figure}[!htbp]
\centering
\includegraphics[width=\linewidth]{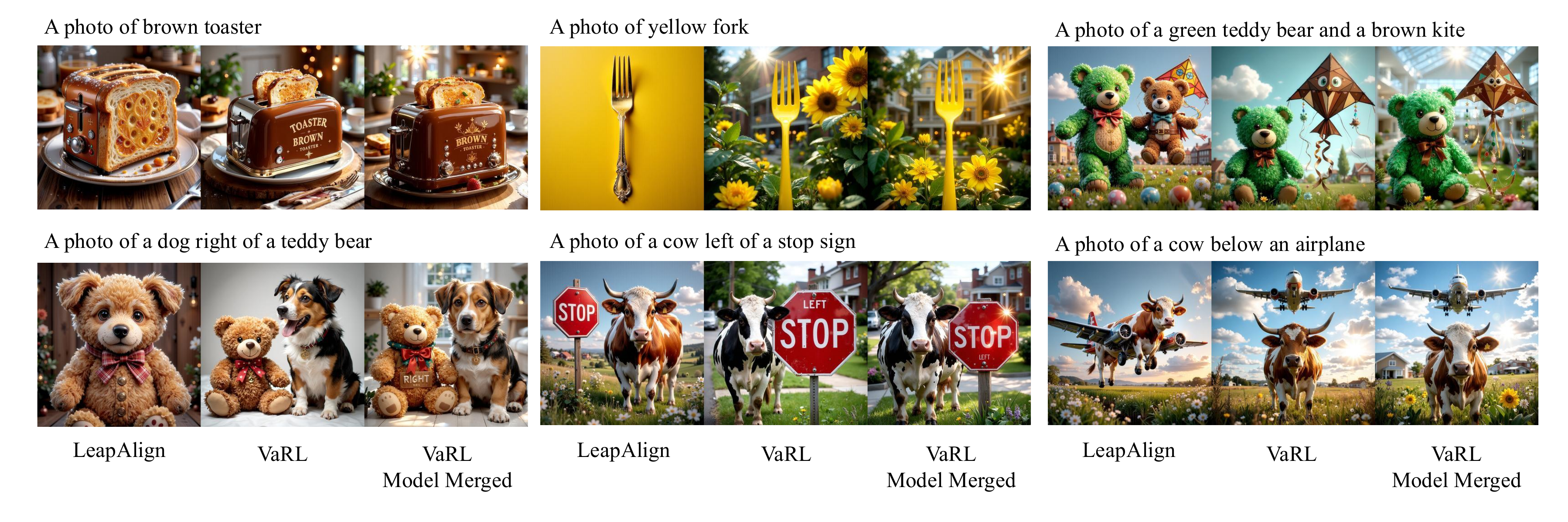}
\caption{
\textbf{Qualitative comparison of compositional alignment on FLUX.1-dev.}
All methods share the same prompt and initial noise; the merged column evaluates the 50/50 checkpoint at 8 steps, demonstrating stronger adherence to specified colors and spatial relations than LeapAlign.
}
\label{fig:flux_color_spatial_demo}
\end{figure}

\begin{table}[!htbp]
\centering
\caption{
FLUX.1-dev performance on the internal rubric-based text-alignment benchmark scored by SeedVL~\citep{seed15vl2025,guo2026editr1} (higher is better).
The merged variant runs 8-step inference using the 50/50 teacher--student checkpoint.
}
\label{tab:flux_alignment}
\small
\renewcommand{\arraystretch}{1.05}
\setlength{\tabcolsep}{4pt}
\begin{tabular}{l c c c c c}
\toprule
Metric & FLUX.1-dev & w. \refl{} & w. LeapAlign & w. VaRL & w. VaRL (merged) \\
\midrule
VLM-RM score & 62.12 & 73.88 & 75.52 & \textbf{80.01} & \underline{78.80} \\
\bottomrule
\end{tabular}
\end{table}

\begin{figure}[!htbp]
\centering
\includegraphics[width=\linewidth]{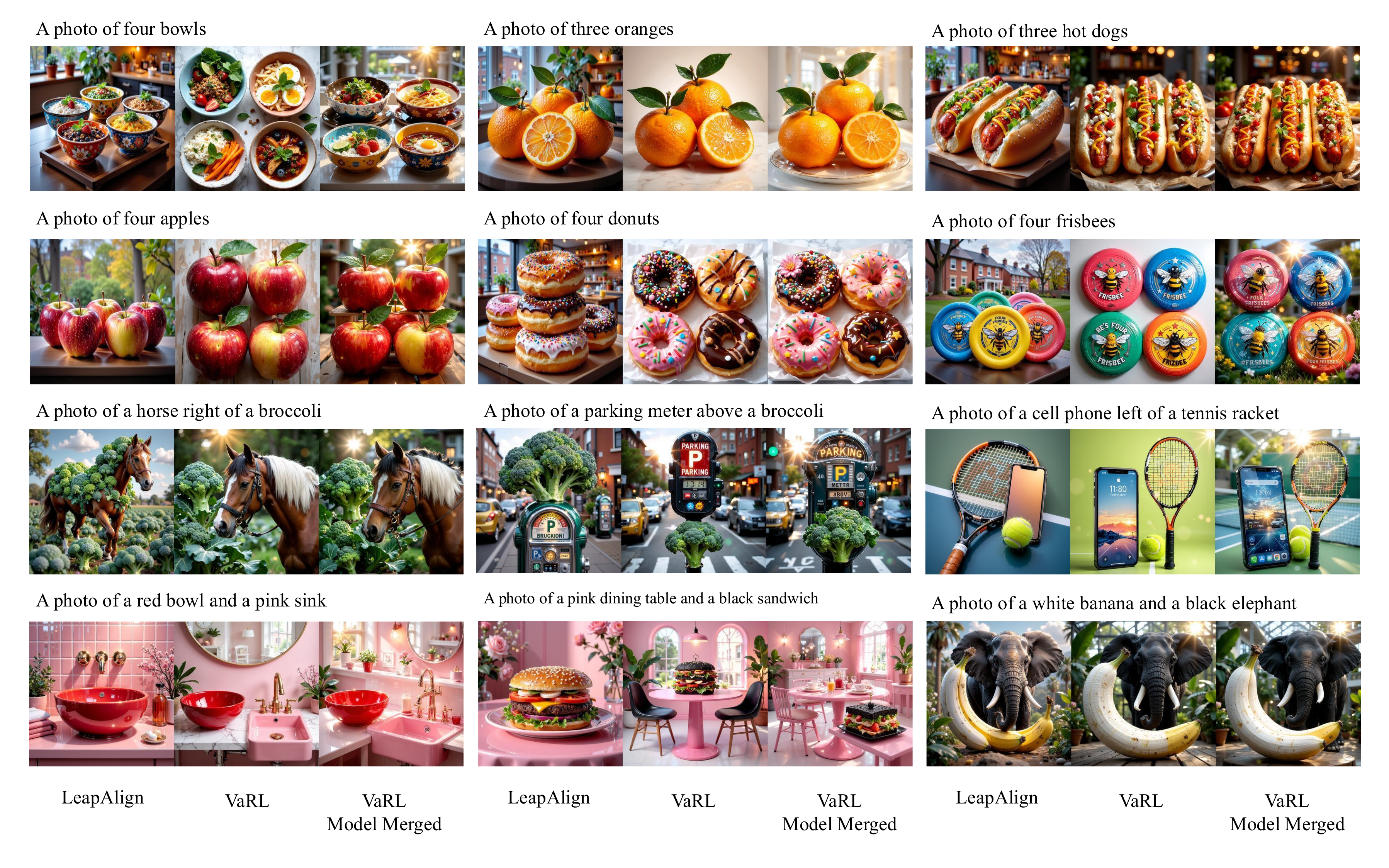}
\caption{
\textbf{Additional qualitative comparison of compositional alignment on FLUX.1-dev.}
All methods share the same prompt and initial noise; the merged column evaluates the 50/50 teacher--student checkpoint at 8 steps.
VaRL improves object counting, spatial positioning, and attribute binding, matching the quantitative gains in Tab.~\ref{tab:hps_geneval_opdo}.
}
\label{fig:flux_count_demo}
\end{figure}

Under 26B VLM-RM optimization (Tab.~\ref{tab:flux_alignment}), VaRL improves the base FLUX.1-dev rubric score from 62.12 to 80.01, outperforming both \refl{} (73.88) and LeapAlign (75.52), while the 8-step 50/50 merged checkpoint retains a score of 78.80.
Qualitative comparisons under matched prompts and initial noise in Fig.~\ref{fig:flux_color_spatial_demo} and Fig.~\ref{fig:flux_count_demo} corroborate the color, spatial-layout, and counting gains in Tab.~\ref{tab:hps_geneval_opdo}.

\subsection{Wanx-T2V-14B Results}

\begin{table}[!htbp]
\centering
\caption{
Human pairwise GSB preference ratios $(G-B)/300$ on Wanx-T2V-14B (8-step uses the 50/50 merged checkpoint).
}
\label{tab:wanx_gsb}
\small
\setlength{\tabcolsep}{4pt}
\renewcommand{\arraystretch}{1.08}
\begin{tabular}{@{}lcc@{}}
\toprule
Comparison & Aesthetic GSB & Motion quality GSB \\
\midrule
w. VaRL vs. Wanx-T2V-14B & 44.33\% & 33.33\% \\
w. VaRL vs. w. \refl{} & 30.33\% & 20.67\% \\
w. VaRL (8-step, merged) vs. w. \refl{} & 23.67\% & 17.33\% \\
\bottomrule
\end{tabular}
\end{table}

\begin{figure}[!htbp]
\centering
\includegraphics[width=\linewidth]{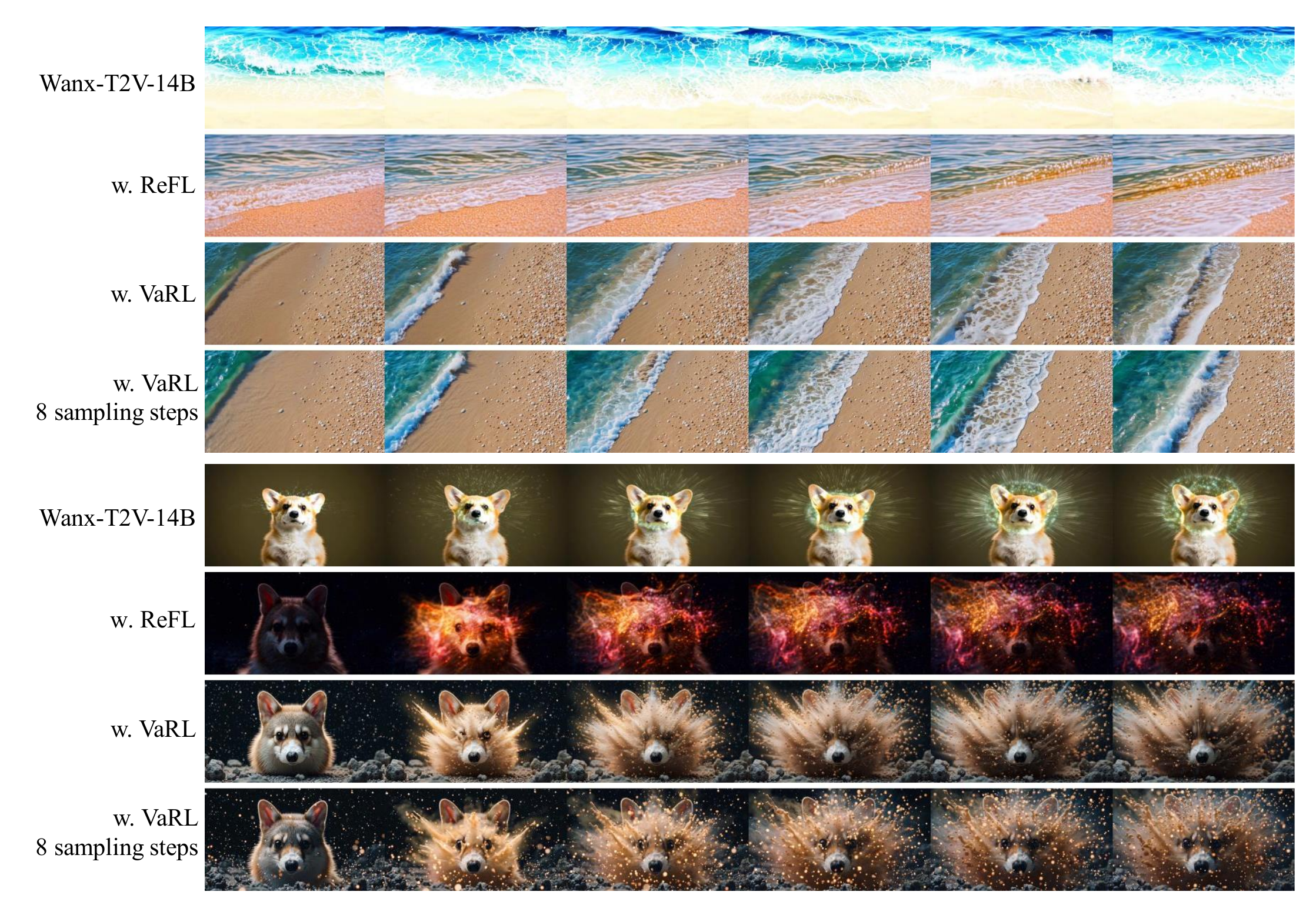}
\caption{
\textbf{Qualitative comparison on Wanx-T2V-14B.}
Uniformly sampled frames from shared initial noise for two prompts: ocean waves washing onto a sandy shoreline (top) and a corgi shaking off glowing dust particles against a dark backdrop (bottom).
Compared with the overexposed coastline in the base model and the darkened subject in \refl{}, both 20-step and 8-step merged VaRL preserve natural shoreline geometry, crisp fur texture, and coherent particle motion.
}
\label{fig:wanx_main_demo}
\end{figure}

We apply the same alternating distillation-and-reward procedure to Wanx-T2V-14B using complete 20-step CFG teacher rollouts (guidance scale 3.5, flow shift 6, with one conditional and one unconditional pass per step).
While the student is supervised by full video rollouts, the staged reward backward pass decodes at most five frames and backpropagates through at most four uniformly strided latent frames to keep GPU memory bounded.
For human evaluation, 15 raters independently judge anonymized video pairs across 20 VBench prompts (300 judgments per row in Tab.~\ref{tab:wanx_gsb}), scoring aesthetic quality and motion quality separately via $(G-B)/300$.
As shown in Tab.~\ref{tab:wanx_gsb}, both the 20-step VaRL policy and the 8-step 50/50 merged model are preferred over the pretrained Wanx-T2V-14B backbone (+44.33\% aesthetic and +33.33\% motion at 20 steps) as well as the 20-step \refl{}-style baseline (+30.33\% and +20.67\% at 20 steps; +23.67\% and +17.33\% at 8 steps).
Fig.~\ref{fig:wanx_main_demo} shows qualitative frame strips generated under shared prompts and initial noise; more comparisons are provided in Appendix Figs.~\ref{fig:wanx_appendix_aesthetic}--\ref{fig:wanx_vis5}.

\subsection{Ablation on Auxiliary Student Losses}
\label{sec:flux_auxiliary_ablation}

Tab.~\ref{tab:flux_auxiliary_ablation} ablates the auxiliary perceptual ($\mathcal{L}_{\mathrm{LPIPS}}$) and consistency ($\mathcal{L}_{\mathrm{LCM}}$) student losses on FLUX.1-dev under HPSv2.1 while keeping the core endpoint and chord-velocity losses ($\mathcal{L}_{x_0}$ and $\mathcal{L}_v$) active.
Without either auxiliary loss (\emph{None}), VaRL already reaches $0.4140$ on HPSv2.1---surpassing LeapAlign ($0.4092$ in Tab.~\ref{tab:hps_geneval_opdo})---confirming that the primary gain comes from rollout-consistent endpoint distillation rather than auxiliary regularizers, consistent with our zero-auxiliary setup on Wanx-T2V-14B.
Adding $\mathcal{L}_{\mathrm{LCM}}$ ($0.4192$) or $\mathcal{L}_{\mathrm{LPIPS}}$ ($0.4188$) individually further sharpens student endpoint predictions, and combining both achieves the best score ($0.4229$).

\begin{table}[!htbp]
\centering
\caption{
\textbf{Ablation of auxiliary student losses on FLUX.1-dev (HPSv2.1).}
Core losses $\mathcal{L}_{x_0}$ and $\mathcal{L}_v$ are enabled in all settings.
}
\label{tab:flux_auxiliary_ablation}
\setlength{\tabcolsep}{8pt}
\renewcommand{\arraystretch}{1.05}
\begin{tabular}{@{}lcccc@{}}
\toprule
 & \textbf{None} & \textbf{LCM only} & \textbf{LPIPS only} & \textbf{LPIPS + LCM} \\
\midrule
HPSv2.1$\uparrow$ & 0.4140 & 0.4192 & 0.4188 & \textbf{0.4229} \\
\bottomrule
\end{tabular}
\end{table}

%% file: sections/conclusion.tex
\FloatBarrier
\section{Conclusion}

We presented VaRL, an on-policy endpoint-distillation framework for diffusion and flow-matching post-training.
By supervising an online student velocity model on complete stop-gradient teacher rollouts and backpropagating differentiable rewards through one local denoising step composed with the student endpoint map, VaRL keeps gradient depth short while anchoring updates to full inference trajectories.
Under local endpoint-accuracy and smoothness conditions, each sampled high-noise update closely approximates that step's full-rollout reward gradient contribution.
On FLUX.1-dev, VaRL achieves 0.4229 HPSv2.1 and 78.58 GenEval with particularly strong gains on counting, spatial layout, and attribute binding over more than 10,000 stable updates; on Wanx-T2V-14B, it reaches +44.33\% aesthetic and +33.33\% motion GSB gains over the pretrained backbone while outperforming a ReFL-style baseline, and simple 50/50 teacher--student weight averaging yields competitive 4- and 8-step generators without extra distillation.

%% file: sections/appendix.tex
\section{Additional Details and Results}

\subsection{Gradient Alignment Proof}
\label{app:gradient_alignment_proof}

We justify Proposition~1 using the continuous-time adjoint formulation of probability-flow ODE samplers, under which discrete-step updates serve as numerical quadrature and Monte Carlo estimates of the continuous adjoint integral.
For clarity, we derive the bound in the synchronized limit $\bar{\theta}=\theta$, where the rollout teacher coincides with the updated policy; in alternating training, frequent teacher refreshes keep any additional lag bounded by single-step parameter drift.

\paragraph{Full rollout gradient.}
Consider the deterministic probability-flow sampler
\begin{equation}
    \frac{d\latent_t}{dt}=v_{\theta}(\latent_t,t,\prompt),
    \qquad t\in[0,T],
\end{equation}
with decoded terminal reward $J(\theta)=\rewardmodel(\decoder(\latent_0),\prompt)$.
Writing $\bar{\rewardmodel}(\latent_0,\prompt)=\rewardmodel(\decoder(\latent_0),\prompt)$ and letting $\Phi_{\theta}^{t\rightarrow 0}(\latent_t,\prompt)$ denote the ODE solution map from time $t$ to $0$, the reward adjoint at time $t$ is
\begin{equation}
    a(t)
    =
    \left(\nabla_{\latent}\Phi_{\theta}^{t\rightarrow 0}(\latent_t,\prompt)\right)^{\top}
    \nabla_{\latent_0}\bar{\rewardmodel}(\latent_0,\prompt).
\end{equation}
By the continuous adjoint method, the full rollout gradient is
\begin{equation}
    \nabla_{\theta}J(\theta)
    =
    \int_{T}^{0}
    \left(\partial_{\theta}v_{\theta}(\latent_t,t,\prompt)\right)^{\top}
    a(t)\,dt .
    \label{eq:continuous_full_gradient}
\end{equation}
Later denoising steps do not introduce separate terms; their effect is carried by $a(t)$ and accumulated over time.
Because generation runs from $T$ down to $0$, the integration order fixes the sign (e.g., for $v_{\theta}=\theta$ and $J=x_0$, $x_0=x_T-T\theta$ gives $\nabla_{\theta}J=-T$, matching Eq.~\eqref{eq:continuous_full_gradient}).

\paragraph{CFG rollout and conditional training.}
In implementation, teacher rollouts follow the backbone's deterministic inference schedule under classifier-free guidance (CFG), evaluating one conditional pass $v_{\bar{\theta}}(\latent_t,t,\prompt)$ and one unconditional pass $v_{\bar{\theta}}(\latent_t,t,\varnothing)$ per step.
Consequently, the rollout endpoint $x_0^{\mathrm{ref}}$ reflects the guided trajectory, and the student $F_{\phi}(\latent_t,t,\prompt)$ distills this CFG-guided dynamics directly into a single conditional pass (allowing merged few-step inference near guidance scale $1$).
During policy optimization (Sec.~\ref{sec:policy_update}), the local step $\latent_s^{\theta}=\mathcal{S}_{s,t}(\latent_t,\hat{v}_{\theta})$ evaluates and updates only the conditional branch $\hat{v}_{\theta}=v_{\theta}(\latent_t,t,\prompt)$, avoiding extra unconditional backward passes while routing reward gradients directly into the prompt-conditioned velocity field.

\paragraph{Student adjoint.}
The student velocity model $F_{\phi}$ defines the composed endpoint map
\begin{equation}
    \studentmap(\latent_t,t,\prompt)
    =
    \mathcal{S}_{0,t}\!\left(
    \latent_t,
    F_{\phi}(\latent_t,t,\prompt)
    \right),
\end{equation}
which approximates $\Phi_{\theta}^{t\rightarrow 0}$ along on-policy states.
The student-induced adjoint and gradient integrand at time $t$ are
\begin{align}
    \hat{a}_{\phi}(t)
    &=
    \left(\nabla_{\latent}\studentmap(\latent_t,t,\prompt)\right)^{\top}
    \nabla_{\latent_0}\bar{\rewardmodel}(\studentmap(\latent_t,t,\prompt),\prompt), \\
    g_{\mathrm{varl}}(t)
    &=
    -
    \left(\partial_{\theta}v_{\theta}(\latent_t,t,\prompt)\right)^{\top}
    \hat{a}_{\phi}(t).
\end{align}

\paragraph{Adjoint alignment bound.}
Assume that at the sampled on-policy state,
\begin{equation}
    \|\studentmap(\latent_t,t,\prompt)-\Phi_{\theta}^{t\rightarrow 0}(\latent_t,\prompt)\|_2
    \leq \epsilon_0.
    \label{eq:endpoint_error}
\end{equation}
Here $\epsilon_0$ combines the student fitting residual and the finite-step solver error.
Replacing the continuous map $\Phi_{\theta}^{t\rightarrow 0}$ with a $q$-th order discrete solver $\Phi_{\theta,h}^{t\rightarrow 0}$ of maximum step size $h$ incurs a flow-map error
\begin{equation}
    \|\Phi_{\theta,h}^{t\rightarrow 0}(\latent_t,\prompt)
    -
    \Phi_{\theta}^{t\rightarrow 0}(\latent_t,\prompt)\|_2
    =
    O(h^q),
\end{equation}
whenever $v_{\theta}$ is Lipschitz in $\latent$~\citep{gholami2019anode,onken2020discretize,domingoenrich2025adjoint}; we absorb this solver gap into $\epsilon_0$.

In practice, policy updates sample $t$ from the high-noise schedule subset rather than uniformly over $[0,T]$, since near $t=0$ one-step endpoint predictions are already accurate.
Even in this high-noise regime, the endpoint prediction error remains well-controlled for two reasons: first, decoded clean images lie in $[-1,1]$ and clean VAE latents $x_0$ occupy a bounded region of latent space, which keeps the conditional endpoint variance bounded; second, sampled discrete indices stay strictly away from the pure-noise singularity or degenerate endpoints (e.g., $k\in[1,16)$ on the 28-step FLUX.1-dev grid and $k\in[3,7]$ on the 20-step Wanx-T2V-14B grid; Tabs.~\ref{tab:flux_training_params} and~\ref{tab:wanx_training_params}), preventing $1/t$ blow-up in the chord-velocity target $v_t^{\mathrm{ref}}$ and keeping $\studentmap$ well-conditioned at every evaluated state.

Assume further that
$\|\nabla_{\latent}\Phi_{\theta}^{t\rightarrow 0}\|_{\mathrm{op}}\leq B_{\Phi}$,
$\|\partial_{\theta}v_{\theta}\|_{\mathrm{op}}\leq B_v$, and $\bar{\rewardmodel}$ has bounded, Lipschitz gradients:
\begin{align}
    \|\nabla_x\bar{\rewardmodel}(x,\prompt)\|_2 &\leq G_R, \nonumber \\
    \|\nabla_x\bar{\rewardmodel}(x,\prompt)-\nabla_x\bar{\rewardmodel}(y,\prompt)\|_2
    &\leq L_{\nabla R}\|x-y\|_2 .
\end{align}

\paragraph{Endpoint fitting controls the first-order term.}
Endpoint regression ($\mathcal{L}_{x_0}$) directly reduces $\epsilon_0$.
To bound the Jacobian discrepancy between $\studentmap$ and $\Phi_{\theta}^{t\rightarrow 0}$, we use a standard finite-difference argument.
For locally Lipschitz ODE vector fields, $\Phi_{\theta}^{t\rightarrow 0}$ is locally Lipschitz in its initial state~\citep{chen2024diffusion}: on $\mathcal{B}_{\rho}(\latent_t)=\{\latent:\|\latent-\latent_t\|_2\leq\rho\}$,
\begin{equation}
    \|\Phi_{\theta}^{t\rightarrow 0}(\latent,\prompt)
    -
    \Phi_{\theta}^{t\rightarrow 0}(\latent',\prompt)\|_2
    \leq
    L_{\Phi}\|\latent-\latent'\|_2 .
\end{equation}
Define the endpoint residual
$e_{\phi}(\latent,t)=\studentmap(\latent,t,\prompt)-\Phi_{\theta}^{t\rightarrow 0}(\latent,\prompt)$.
Suppose on-policy distillation bounds this error uniformly on $\mathcal{B}_{\rho}(\latent_t)$,
\begin{equation}
    \sup_{\latent\in\mathcal{B}_{\rho}(\latent_t)}
    \|e_{\phi}(\latent,t)\|_2
    \leq
    \epsilon_{\mathrm{loc}},
\end{equation}
and $e_{\phi}(\cdot,t)$ has $\beta_e$-Lipschitz gradient. For any unit vector $u$ and $0<\eta\leq\rho$,
\begin{equation}
    e_{\phi}(\latent_t+\eta u,t)-e_{\phi}(\latent_t,t)
    =
    \int_0^{\eta}
    \nabla_{\latent}e_{\phi}(\latent_t+\tau u,t)u\,d\tau .
\end{equation}
Subtracting $\eta\nabla_{\latent}e_{\phi}(\latent_t,t)u$ and applying $\beta_e$-smoothness yields
\begin{equation}
    \left\|
    \frac{e_{\phi}(\latent_t+\eta u,t)-e_{\phi}(\latent_t,t)}{\eta}
    -
    \nabla_{\latent}e_{\phi}(\latent_t,t)u
    \right\|_2
    \leq
    \frac{\beta_e \eta}{2},
\end{equation}
and hence, taking the supremum over unit vectors $u$,
\begin{equation}
    \|\nabla_{\latent}\studentmap
    -
    \nabla_{\latent}\Phi_{\theta}^{t\rightarrow 0}\|_{\mathrm{op}}
    =
    \|\nabla_{\latent}e_{\phi}\|_{\mathrm{op}}
    \leq
    \frac{2\epsilon_{\mathrm{loc}}}{\eta}
    +
    \frac{\beta_e \eta}{2}.
\end{equation}
Minimizing over $\eta\in(0,\rho]$ defines
\begin{equation}
    \delta_{\nabla}
    :=
    \inf_{0<\eta\leq\rho}
    \left(
    \frac{2\epsilon_{\mathrm{loc}}}{\eta}
    +
    \frac{\beta_e \eta}{2}
    \right),
\end{equation}
which gives $\delta_{\nabla}=\sqrt{4\beta_e\epsilon_{\mathrm{loc}}}=O(\sqrt{\beta_e\epsilon_{\mathrm{loc}}})$ whenever $\sqrt{4\epsilon_{\mathrm{loc}}/\beta_e}\leq\rho$ (and implies the local Lipschitz bound $L_{\phi}\leq L_{\Phi}+\delta_{\nabla}$ on $\studentmap$).
Auxiliary consistency or perceptual losses~\citep{song2023consistency,luo2023latent} (used on FLUX.1-dev, disabled on Wanx-T2V-14B) empirically promote local smoothness, though the bound requires only $\epsilon_{\mathrm{loc}}$ and $\beta_e$.

Abbreviating $G=\studentmap(\latent_t,t,\prompt)$ and $\Phi=\Phi_{\theta}^{t\rightarrow 0}(\latent_t,\prompt)$, the triangle inequality gives
\begin{align}
    \|\hat{a}_{\phi}(t)-a(t)\|_2
    &\leq
    \left\|
    \left(\nabla_{\latent}\studentmap
    -
    \nabla_{\latent}\Phi_{\theta}^{t\rightarrow 0}\right)^{\top}
    \nabla_x\bar{\rewardmodel}(G,\prompt)
    \right\|_2 \nonumber \\
    &+
    \bigl\|
    \bigl(\nabla_{\latent}\Phi_{\theta}^{t\rightarrow 0}\bigr)^{\!\top}\!
    \bigl[
    \nabla_x\bar{\rewardmodel}(G,\prompt)
    -
    \nabla_x\bar{\rewardmodel}(\Phi,\prompt)
    \bigr]
    \bigr\|_2 \nonumber \\
    &\leq
    G_R\delta_{\nabla}+B_{\Phi}L_{\nabla R}\epsilon_0 .
    \label{eq:continuous_adjoint_error}
\end{align}
Writing $g_{\mathrm{full}}(t)=-(\partial_{\theta}v_{\theta}(\latent_t,t,\prompt))^{\top}a(t)$ so that $\nabla_{\theta}J=\int_0^T g_{\mathrm{full}}(t)\,dt$ in Eq.~\eqref{eq:continuous_full_gradient}, we obtain
\begin{equation}
    \|g_{\mathrm{varl}}(t)
    -
    g_{\mathrm{full}}(t)\|_2
    \leq
    B_v\left(G_R\delta_{\nabla}+B_{\Phi}L_{\nabla R}\epsilon_0\right).
    \label{eq:continuous_integrand_error}
\end{equation}

\paragraph{From sampled timesteps to the full gradient.}
Under uniform sampling $t\sim\mathcal{U}[0,T]$, integrating Eq.~\eqref{eq:continuous_integrand_error} over $[0,T]$ yields
\begin{align}
    \left\|
    T\,\expect_t[g_{\mathrm{varl}}(t)]
    -
    \nabla_{\theta}J(\theta)
    \right\|_2
    &\leq
    \int_0^T
    \|g_{\mathrm{varl}}(t)-g_{\mathrm{full}}(t)\|_2\,dt \nonumber \\
    &\leq
    T B_v\left(G_R\delta_{\nabla}+B_{\Phi}L_{\nabla R}\epsilon_0\right),
\end{align}
recovering the $O(\epsilon_0+\sqrt{\epsilon_{\mathrm{loc}}})$ bound in Proposition~1.

\paragraph{Discrete one-step update on non-uniform schedules.}
In practice (Sec.~\ref{sec:policy_update}), sampling runs on a discrete grid $T=t_N > \cdots > t_0=0$ (including non-uniform or flow-shifted schedules) with step sizes $\Delta t_k = t_k - t_{k-1} > 0$ and $h=\max_k \Delta t_k$.
Because $\studentmap$ does not depend on $\theta$, the policy takes one Euler step $\latent_{t_{k-1}}^{\theta}=\latent_{t_k}-\Delta t_k v_{\theta}(\latent_{t_k},t_k,\prompt)$ before querying $\studentmap(\latent_{t_{k-1}}^{\theta},t_{k-1},\prompt)$.
At $\bar{\theta}=\theta$, the post-step state $\latent_{t_{k-1}}^{\theta}=\latent_{t_{k-1}}$ lies on the rollout trajectory, and the chain rule gives the step-$k$ gradient
\begin{equation}
    \nabla_{\theta}\bar{\rewardmodel}(\hat{x}_0,\prompt)
    =
    -\Delta t_k\left(\partial_{\theta}v_{\theta}(\latent_{t_k},t_k,\prompt)\right)^{\top}\hat{a}_{\phi}(t_{k-1}).
\end{equation}
The Euler step Jacobian $-\Delta t_k I$ thus provides the exact Riemann quadrature weight $\Delta t_k$ for $\int_0^T g_{\mathrm{full}}(t)\,dt$ on non-uniform grids, while evaluating $\hat{a}_{\phi}$ at $t_{k-1}$ rather than $t_k$ incurs only an $O(\Delta t_k)$ shift under Lipschitz continuity along the trajectory.
Summing over $k$ (or rescaling a sampled step) therefore matches $\nabla_{\theta}J(\theta)$ up to $O(\epsilon_0+\sqrt{\epsilon_{\mathrm{loc}}}+h)$.

\paragraph{Scope of the bound.}
Proposition~1 bounds the discrepancy between the student-induced update and the exact full-rollout gradient of the differentiable training reward $\rewardmodel$ along visited trajectories; generalization to held-out reward models and human raters is evaluated empirically in Sec.~\ref{sec:experiments}.

\subsection{Notation}

Tab.~\ref{tab:notation_summary} lists the main symbols used in Sec.~\ref{sec:method} and Appendix~\ref{app:gradient_alignment_proof}.

\begin{table}[!htbp]
\centering
\caption{Summary of notation in the main text and appendix.}
\label{tab:notation_summary}
\footnotesize
\setlength{\tabcolsep}{6pt}
\renewcommand{\arraystretch}{1.08}
\begin{tabular}{@{}l p{0.31\textwidth} @{\hspace{18pt}} l p{0.32\textwidth}@{}}
\toprule
\textbf{Symbol} & \textbf{Description} & \textbf{Symbol} & \textbf{Description} \\
\midrule
$\prompt,\;\latent_t,\;\latent_0$ & Prompt, noisy latent at $t$, and clean terminal latent ($t\in[0,T]$ from clean data to noise) & $\Phi_{\theta}^{t\rightarrow 0},\;\Phi_{\theta,h}^{t\rightarrow 0}$ & Continuous ODE flow map ($t\to 0$) and discrete solver map (max step $h$) \\
$I_0 = \decoder(\latent_0)$ & Decoded image or video via latent decoder $\decoder$ & $x_0^{\mathrm{ref}},\;v_t^{\mathrm{ref}}$ & Rollout endpoint and chord velocity $(\latent_t-x_0^{\mathrm{ref}})/t$ \\
$\rewardmodel,\;\widetilde{\rewardmodel},\;\bar{\rewardmodel}$ & Reward model, clipped reward, and $\rewardmodel(\decoder(\cdot),\prompt)$ & $V_{\phi}$ & Student value proxy $\rewardmodel(\decoder(\studentmap(\latent_t,t,\prompt)),\prompt)$ \\
$\diffpolicy,\;\pi_{\bar{\theta}},\;\theta_{\mathrm{ref}}$ & Policy ($\theta$), rollout teacher ($\bar{\theta}$), and reference weights & $a(t),\;\hat{a}_{\phi}(t)$ & Exact rollout adjoint and student-induced adjoint \\
$v_{\theta},\;F_{\phi}$ & Policy velocity field and student velocity network & $\epsilon_0,\;\epsilon_{\mathrm{loc}}$ & Pointwise error at $\latent_t$ and uniform error on $\mathcal{B}_{\rho}(\latent_t)$ \\
$\mathcal{S}_{s,t},\;\mathcal{S}_{0,t}$ & Euler step $\latent_t-(t-s)v$ and endpoint map $\latent_t-tv$ & $\beta_e,\;\delta_{\nabla}$ & Smoothness of $e_{\phi}$ and Jacobian gap $O(\sqrt{\beta_e\epsilon_{\mathrm{loc}}})$ \\
$\studentmap$ & Student endpoint map $\mathcal{S}_{0,t}(\latent_t,F_{\phi}(\latent_t,t,\prompt))$ & $B_{\Phi},\;B_v,\;G_R,\;L_{\nabla R}$ & Bounds on $\|\nabla_{\latent}\Phi_{\theta}^{t\rightarrow 0}\|_{\mathrm{op}}$, $\|\partial_{\theta}v_{\theta}\|_{\mathrm{op}}$, $\|\nabla\bar{\rewardmodel}\|_2$, $\mathrm{Lip}(\nabla\bar{\rewardmodel})$ \\
\bottomrule
\end{tabular}
\end{table}

\subsection{Training Hyperparameters for Different Backbones}
\label{app:training_hyperparameters}

Tabs.~\ref{tab:flux_training_params} and \ref{tab:wanx_training_params} summarize the training setups for FLUX.1-dev and Wanx-T2V-14B.
Discrete scheduler indices $k \in \{1,\ldots,N\}$ run from pure noise ($k=1$) to clean data ($k=N$), opposite to continuous time $t\in[0,T]$ ($t=T$ at noise, $t=0$ at data); small indices $k$ thus denote high-noise steps on each backbone's native schedule.

\begin{table}[!htbp]
\centering
\caption{
\textbf{FLUX.1-dev post-training hyperparameters.}
Teacher rollouts use the standard 28-step deterministic FLUX.1-dev schedule; high-noise states are sampled along each rollout for student distillation and local policy updates.
}
\label{tab:flux_training_params}
\footnotesize
\setlength{\tabcolsep}{4pt}
\renewcommand{\arraystretch}{0.93}
\begin{tabular}{@{}p{0.16\textwidth} p{0.29\textwidth} p{0.13\textwidth} p{0.36\textwidth}@{}}
\toprule
Category & Hyperparameter & Value & Notes \\
\midrule
Backbone \& rollout & Base model / resolution & FLUX.1-dev, $1024^2$ & Shared initialization for teacher, policy, and student \\
 & Rollout steps ($N$) / CFG scale & $28$ / $4.0$ & Deterministic schedule (conditional + unconditional pass) \\
\midrule
Policy update & Learning rate / batch size & $3\times 10^{-6}$ / $1$ & Per-device rollout batch size; 1 policy step per rollout \\
 & Reward clipping / $\beta_{\mathrm{KL}}$ & Enabled / $0.004$ & Clipped reward; KL regularization to $\theta_{\mathrm{ref}}$ \\
\midrule
Student distillation & Learning rate / update ratio & $5\times 10^{-6}$ / $1{:}1$ & Alternates one student step with one policy step \\
 & Sampled step indices ($k$) & $[1, 16)$ & Noise-end indices on the 28-step grid ($N-k \in (12,27]$) \\
 & Weights $(\lambda_{x_0},\lambda_v,\lambda_{\mathrm{LPIPS}},\lambda_{\mathrm{LCM}})$ & $1.0, 1.0, 1.0, 0.5$ & Endpoint, chord-velocity, perceptual, and consistency losses \\
\bottomrule
\end{tabular}

\vspace{1em}
\caption{
\textbf{Wanx-T2V-14B post-training hyperparameters.}
Training alternates 1:1 between student distillation on 20-step rollouts and staged local policy updates with frame subsampling.
}
\label{tab:wanx_training_params}
\footnotesize
\setlength{\tabcolsep}{4pt}
\renewcommand{\arraystretch}{0.93}
\begin{tabular}{@{}p{0.16\textwidth} p{0.29\textwidth} p{0.13\textwidth} p{0.36\textwidth}@{}}
\toprule
Category & Hyperparameter & Value & Notes \\
\midrule
Backbone \& rollout & Base model / resolution & Wanx-T2V-14B & $480\times 832$ spatial resolution, $81$ frames \\
 & Rollout steps ($N$) / CFG / shift & $20$ / $3.5$ / $6$ & Deterministic flow-shifted video schedule \\
\midrule
Policy update & Learning rate / step indices ($k$) & $1\times 10^{-5}$ / $[3, 7]$ & Conditional branch updated at noise-end indices $[3,7]$ \\
 & Reward clipping / $\beta_{\mathrm{KL}}$ & Enabled / $0$ & Clipped reward; reference regularizer disabled \\
 & Decoded / replay frames & $\le 5$ / $\le 4$ & Max decoded frames for reward and strided latent frames in replay \\
\midrule
Student distillation & Learning rate / update ratio & $1\times 10^{-5}$ / $1{:}1$ & Alternates 1:1 with policy updates (no warmup) \\
 & Distillation step indices ($k$) & $[3, 9]$ & Noise-end indices supervised along the 20-step rollout \\
 & Weights $(\lambda_{x_0},\lambda_v,\lambda_{\mathrm{LPIPS}},\lambda_{\mathrm{LCM}})$ & $1.0, 1.0, 0, 0$ & Core endpoint and velocity losses only; auxiliary losses disabled \\
\bottomrule
\end{tabular}
\end{table}

\FloatBarrier
\subsection{Additional Visualizations}
\label{app:additional_video_visualizations}

\begin{figure}[!htbp]
\centering
\includegraphics[width=0.86\textwidth]{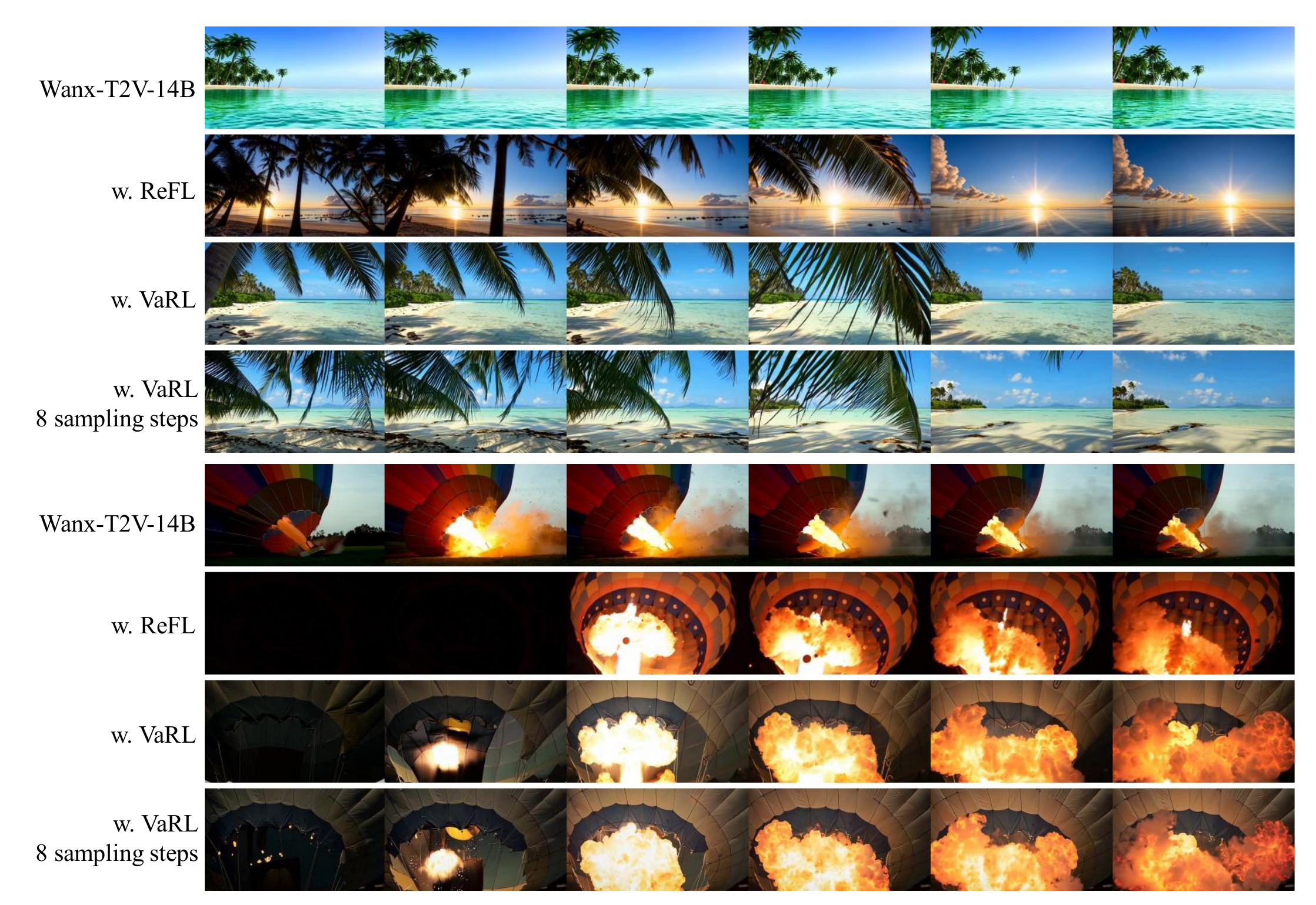}
\caption{
\textbf{Additional Wanx-T2V-14B aesthetic comparisons.}
Uniformly sampled frames from shared initial noise for two prompts: swaying palm fronds framing a clear tropical beach (top) and burner flames illuminating the inside of a hot-air balloon envelope at night (bottom).
VaRL avoids the heavy sunset color cast and early black frames seen in \refl{}, producing balanced daylight exposure and gradual interior illumination.
}
\label{fig:wanx_appendix_aesthetic}
\end{figure}

\begin{figure}[!htbp]
\centering
\includegraphics[width=0.86\textwidth]{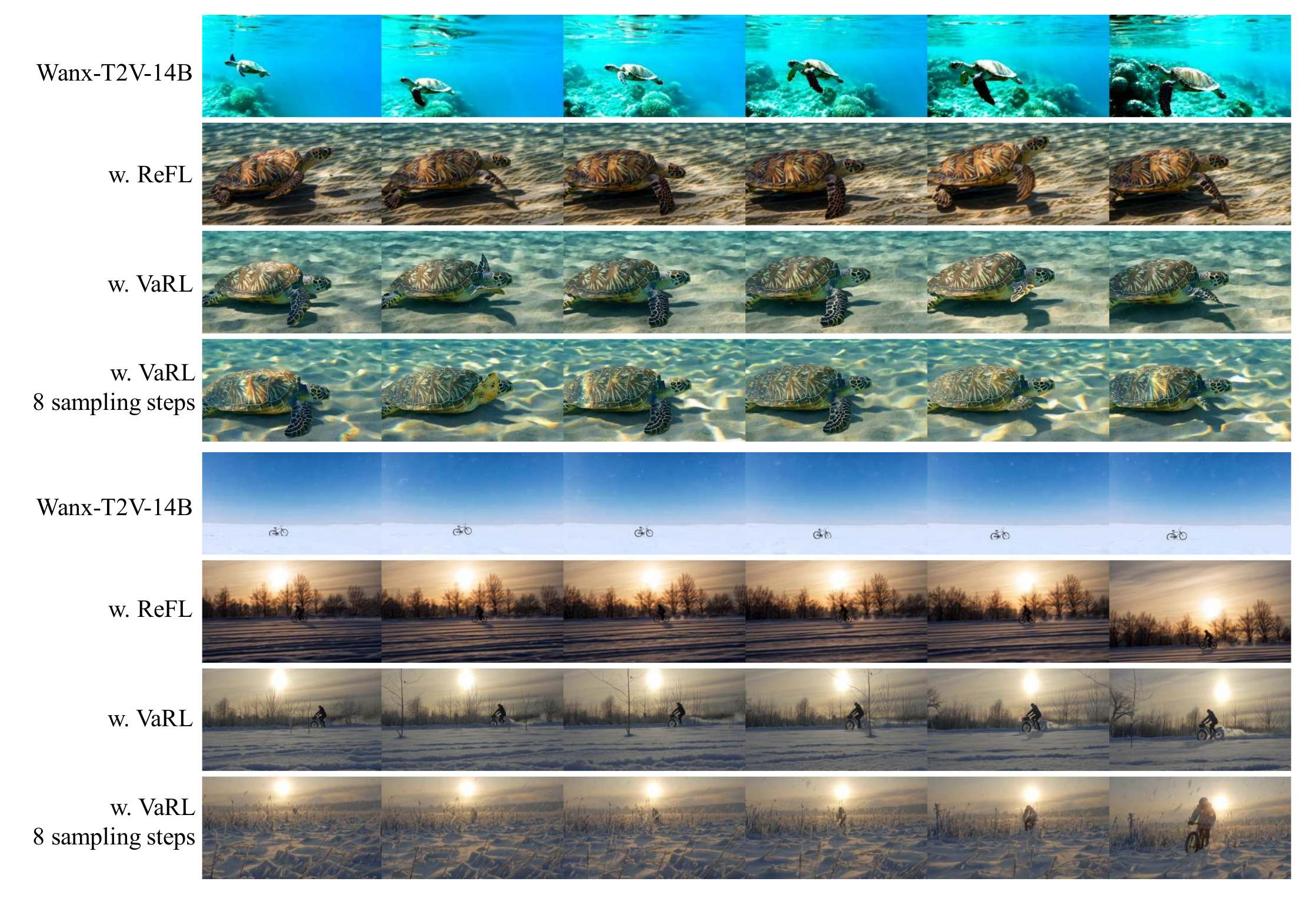}
\caption{
\textbf{Additional Wanx-T2V-14B motion comparisons.}
Uniformly sampled frames from shared initial noise for two prompts: a sea turtle gliding over sunlit shallow sand (top) and a cyclist riding across a snowfield at low sun (bottom).
VaRL maintains consistent shell geometry with realistic water caustics and steady forward cycling motion, whereas the base model loses ground texture in the snow scene.
}
\label{fig:wanx_appendix_motion}
\end{figure}

Extending Fig.~\ref{fig:wanx_main_demo}, Figs.~\ref{fig:wanx_appendix_aesthetic} and \ref{fig:wanx_appendix_motion} compare Wanx-T2V-14B, the \refl{} baseline, our 20-step VaRL policy, and the 8-step 50/50 merged model under identical prompts and initial noise.

\begin{figure}[!htbp]
\centering
\includegraphics[width=\textwidth]{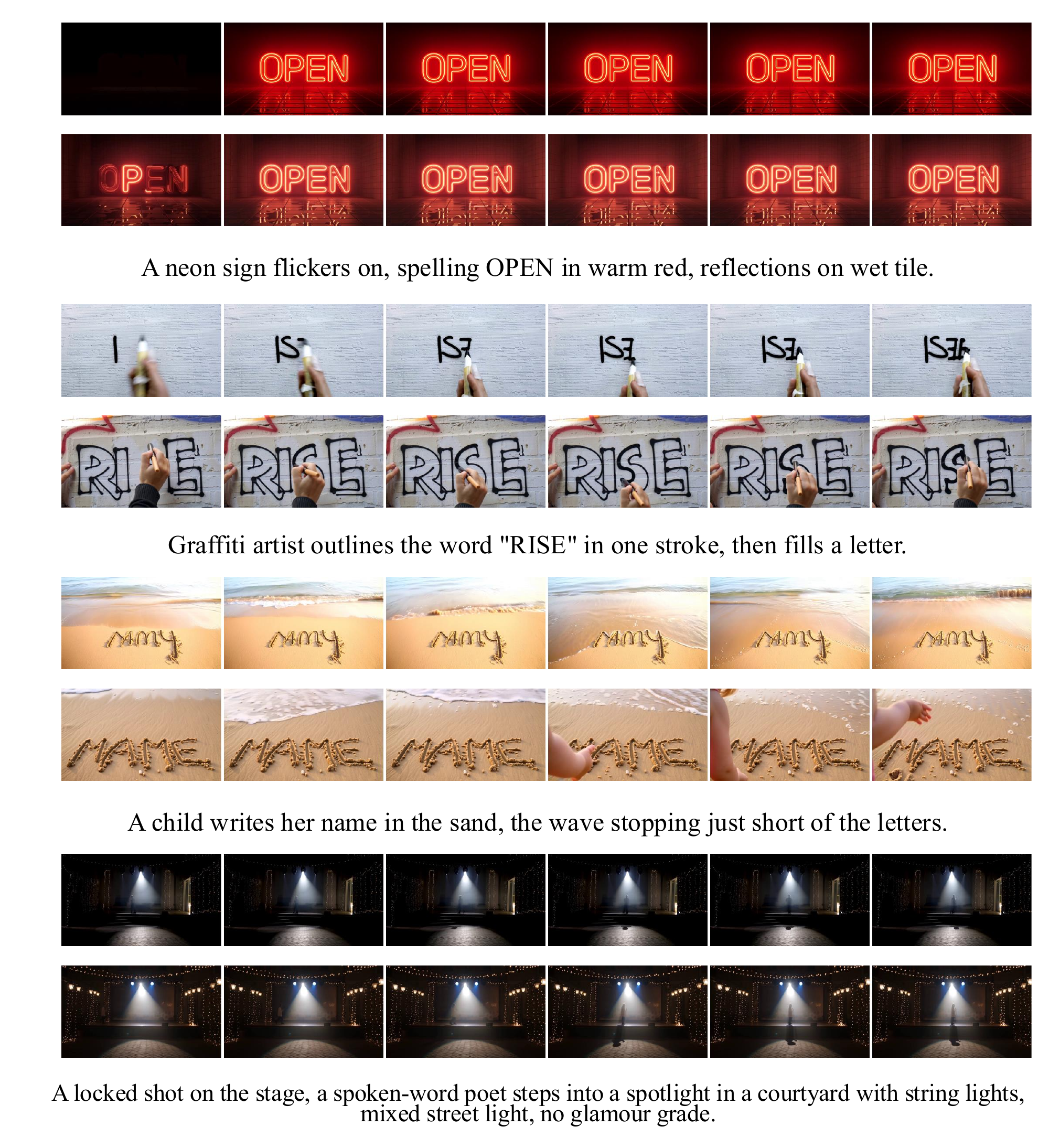}
\caption{
\textbf{Wanx-T2V-14B vs.\ VaRL on scene text and staged motion.}
Six uniformly sampled frames per video from shared prompts and initial noise (upper strip: pretrained Wanx-T2V-14B; lower strip: VaRL) across the \emph{OPEN} neon sign, \emph{RISE} graffiti, sand-lettering, and courtyard-poet scenes.
}
\label{fig:wanx_vis_text_motion}
\end{figure}

\begin{figure}[!htbp]
\centering
\includegraphics[width=\textwidth]{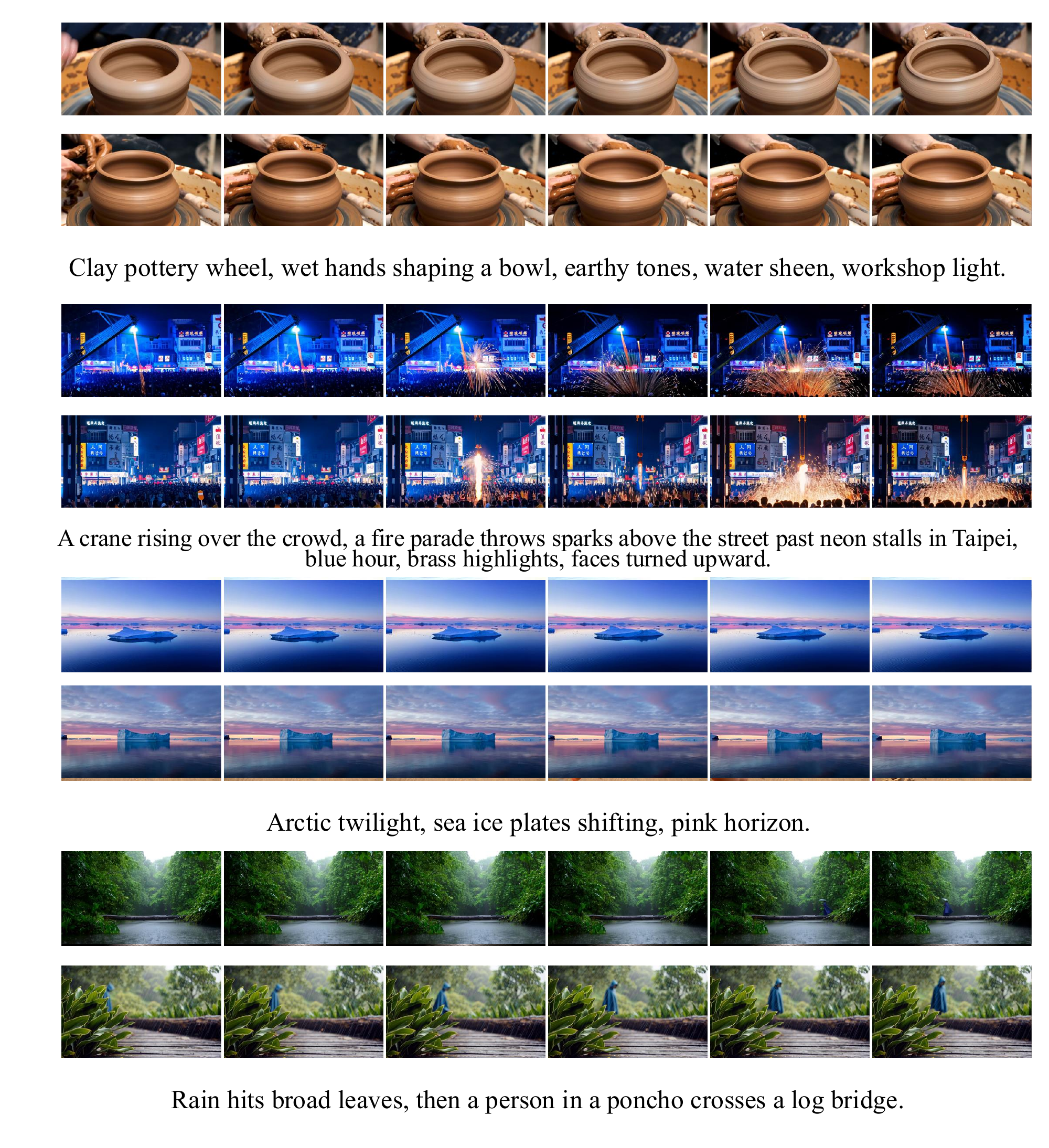}
\caption{
\textbf{Wanx-T2V-14B vs.\ VaRL on material detail and atmospheric lighting.}
Six uniformly sampled frames per video (upper strip: pretrained Wanx-T2V-14B; lower strip: VaRL) across the pottery wheel, Taipei fire parade, Arctic sea ice, and rainy log-bridge prompts.
}
\label{fig:wanx_vis_more}
\end{figure}

\begin{figure}[!htbp]
\centering
\includegraphics[width=\textwidth]{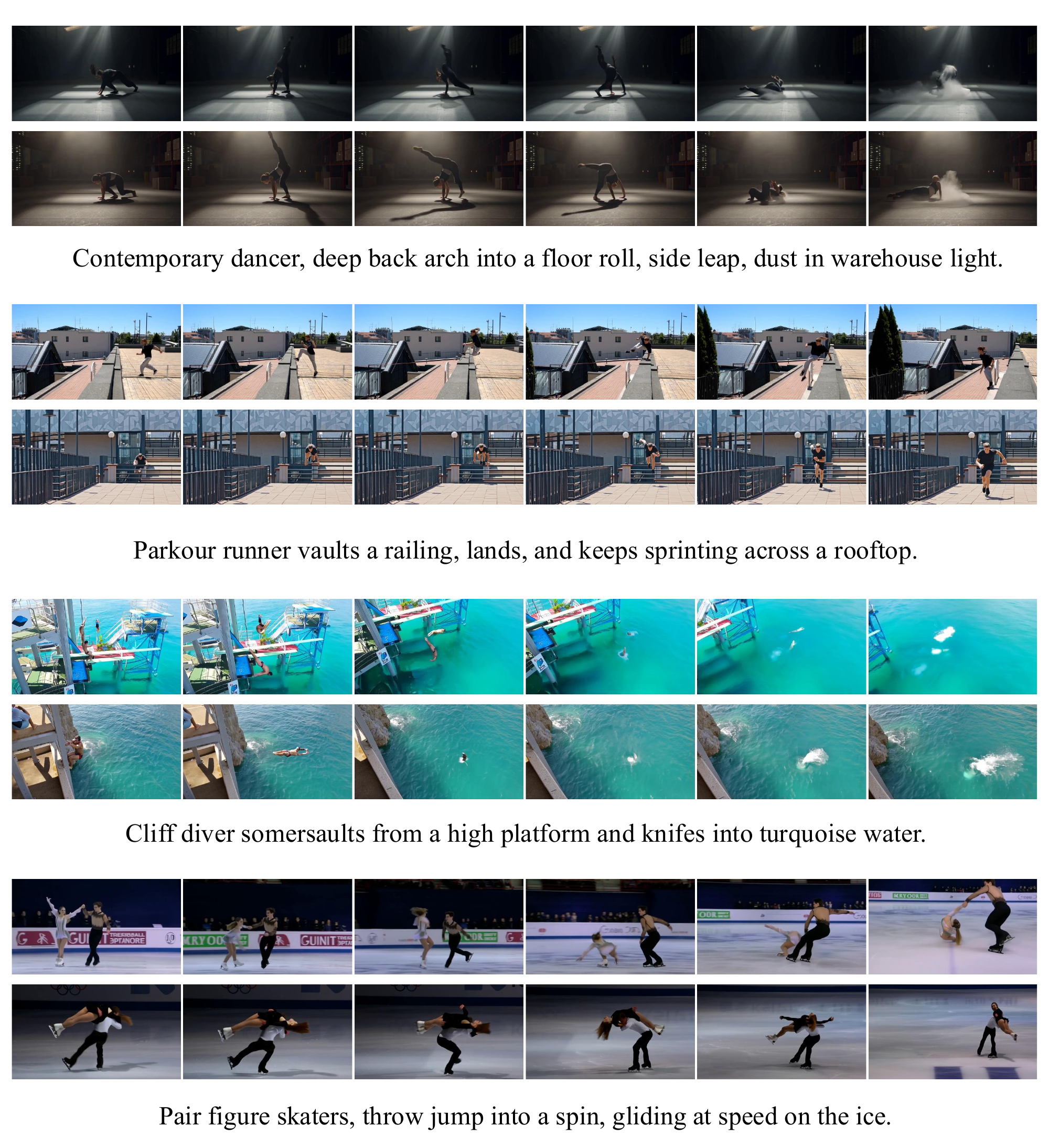}
\caption{
\textbf{Wanx-T2V-14B vs.\ VaRL on complex human motion.}
Six uniformly sampled frames per video (upper strip: pretrained Wanx-T2V-14B; lower strip: VaRL) for contemporary dance, rooftop parkour, cliff diving, and pair figure skating, showing fewer limb artifacts and smoother landings under VaRL.
}
\label{fig:wanx_vis3}
\end{figure}

\begin{figure}[!htbp]
\centering
\includegraphics[width=\textwidth]{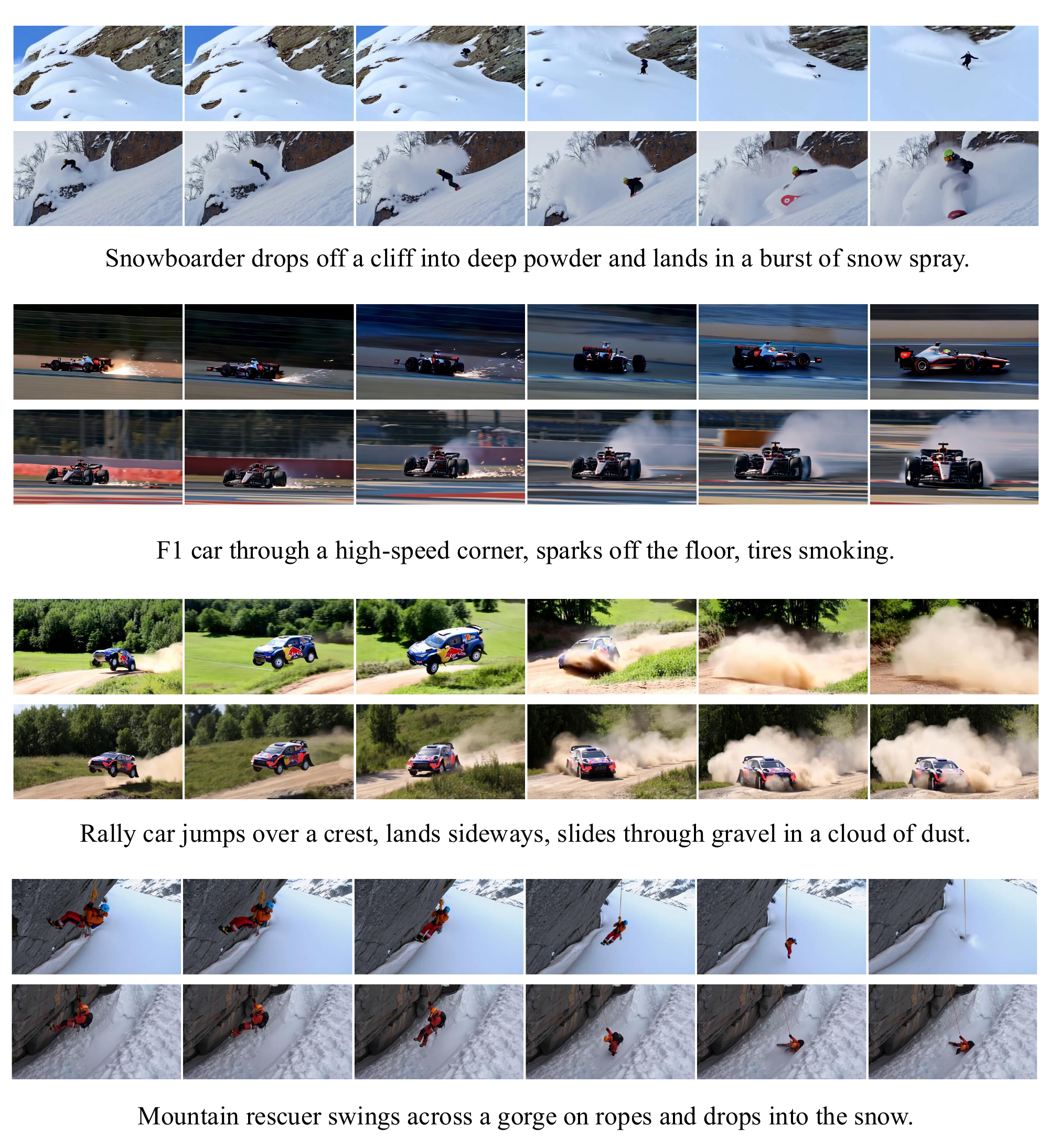}
\caption{
\textbf{Wanx-T2V-14B vs.\ VaRL on high-speed dynamics and particle effects.}
Six uniformly sampled frames per video (upper strip: pretrained Wanx-T2V-14B; lower strip: VaRL) for cliff snowboarding, F1 cornering, rally-car gravel drifting, and rope gorge rescue.
}
\label{fig:wanx_vis4}
\end{figure}

\begin{figure}[!htbp]
\centering
\includegraphics[width=\textwidth]{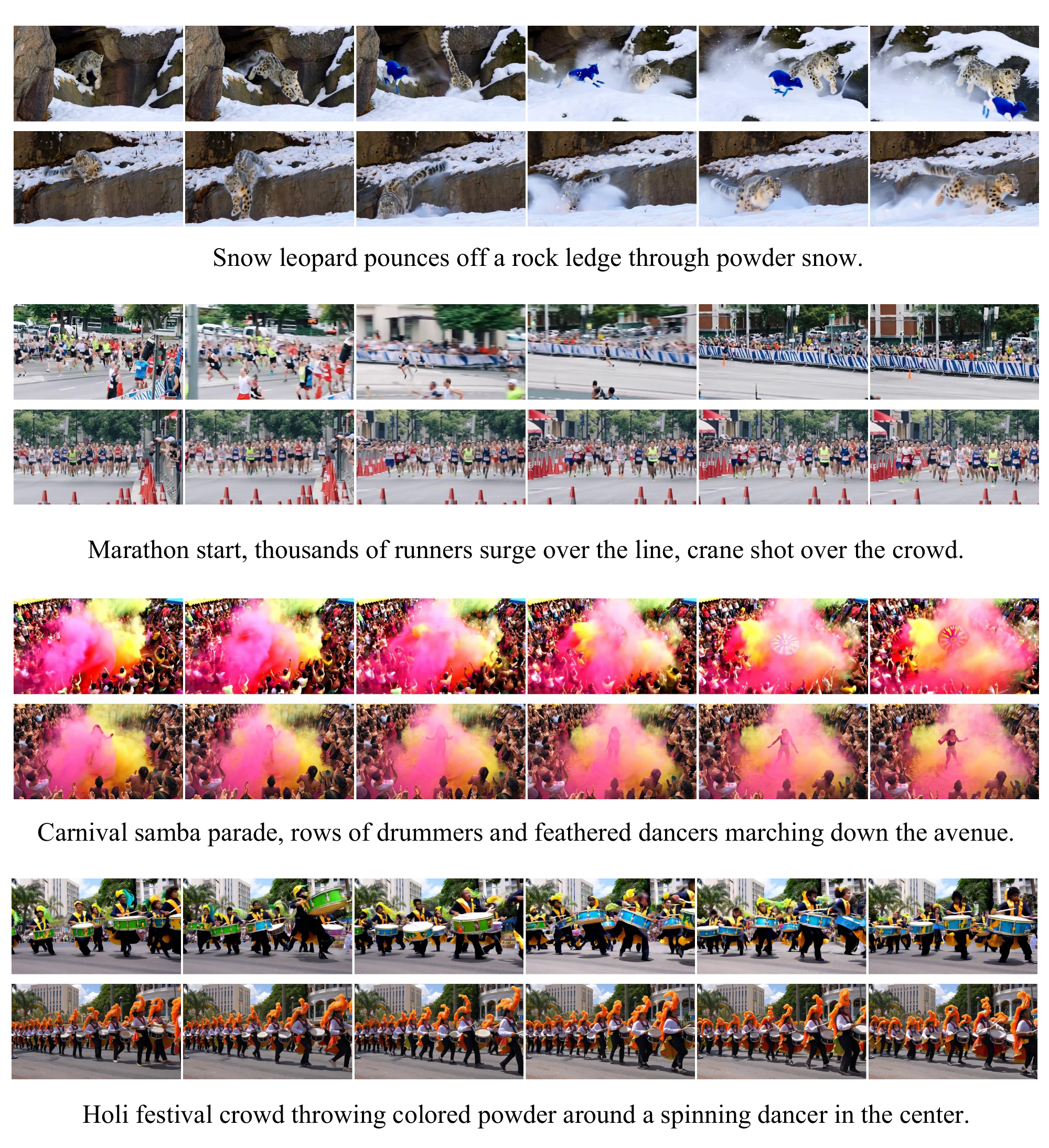}
\caption{
\textbf{Wanx-T2V-14B vs.\ VaRL on wildlife and dense crowd motion.}
Six uniformly sampled frames per video (upper strip: pretrained Wanx-T2V-14B; lower strip: VaRL) across snow-leopard pouncing, marathon start, samba parade, and Holi festival scenes.
}
\label{fig:wanx_vis5}
\end{figure}

Figs.~\ref{fig:wanx_vis_text_motion}--\ref{fig:wanx_vis5} further compare 20-step, 480p generations from pretrained Wanx-T2V-14B (upper strip in each pair) and VaRL (lower strip) under matched prompts and initial noise across five challenging categories:
\begin{itemize}[leftmargin=1.4em,itemsep=2pt,topsep=3pt]
    \item \textbf{Scene text and staged action (Fig.~\ref{fig:wanx_vis_text_motion}).}
    For the \emph{OPEN} neon sign and \emph{RISE} graffiti prompts, the pretrained model misses the initial flicker or produces garbled strokes (``ISZ''), whereas VaRL renders legible lettering, wet-tile reflections, and progressive stroke filling. In the beach-writing and courtyard-poet scenes, VaRL better preserves shoreline interaction and warm string-light ambiance as the speaker steps into the spotlight.
    \item \textbf{Fine material dynamics and atmospheric lighting (Fig.~\ref{fig:wanx_vis_more}).}
    VaRL captures wet clay sheen and steady hand-rim contact on the pottery wheel, upward spark trajectories in the Taipei night parade without crane artifacts, crisp tabular sea-ice edges at twilight, and a clear foreground perspective of the poncho-clad figure crossing the rain-swept log bridge.
    \item \textbf{Acrobatic human motion (Fig.~\ref{fig:wanx_vis3}).}
    Across fast full-body actions---contemporary floor rolls, rooftop parkour vaults, high-platform cliff diving, and pair figure-skating lifts---pretrained Wanx-T2V-14B often exhibits limb distortion, mid-air stalls, or duplicated divers. VaRL maintains articulated body proportions and continuous momentum from takeoff through landing.
    \item \textbf{High-speed vehicles and particle physics (Fig.~\ref{fig:wanx_vis4}).}
    In extreme-sport and motorsport prompts, VaRL produces fuller powder plumes around the cliff-dropping snowboarder, sustained tire smoke and underbody sparks as the F1 car rounds the apex, coherent sideways drift with trailing gravel dust for the jumping rally car, and stable rope tension for the swinging mountain rescuer.
    \item \textbf{Wildlife dynamics and dense crowd scenes (Fig.~\ref{fig:wanx_vis5}).}
    Pretrained Wanx-T2V-14B introduces stray color artifacts during the snow-leopard pounce and loses structural coherence in dense crowds. VaRL removes the spurious blue patch on the snow slope, keeps marathon runners advancing uniformly toward the camera, and preserves organized parade rows and vivid powder bursts across the carnival and Holi festival scenes.
\end{itemize}

%% file: main.bbl
\begin{thebibliography}{68}
\providecommand{\natexlab}[1]{#1}
\providecommand{\url}[1]{\texttt{#1}}
\expandafter\ifx\csname urlstyle\endcsname\relax
  \providecommand{\doi}[1]{doi: #1}\else
  \providecommand{\doi}{doi: \begingroup \urlstyle{rm}\Url}\fi

\bibitem[Black et~al.(2023)Black, Janner, Du, Kostrikov, and Levine]{black2023ddpo}
Kevin Black, Michael Janner, Yilun Du, Ilya Kostrikov, and Sergey Levine.
\newblock Training diffusion models with reinforcement learning.
\newblock \emph{arXiv preprint arXiv:2305.13301}, 2023.

\bibitem[{Black Forest Labs}(2024)]{flux2024}
{Black Forest Labs}.
\newblock {FLUX}.1-dev.
\newblock \url{https://github.com/black-forest-labs/flux}, 2024.

\bibitem[Blattmann et~al.(2023)Blattmann, Dockhorn, Kulal, Mendelevitch, Kilian, Lorenz, Levi, English, Voleti, Letts, et~al.]{blattmann2023alignyourlatents}
Andreas Blattmann, Tim Dockhorn, Sumith Kulal, Daniel Mendelevitch, Maciej Kilian, Dominik Lorenz, Yam Levi, Zion English, Vikram Voleti, Adam Letts, et~al.
\newblock Align your latents: High-resolution video synthesis with latent diffusion models.
\newblock In \emph{Proceedings of the IEEE/CVF Conference on Computer Vision and Pattern Recognition}, 2023.

\bibitem[{ByteDance Seed}(2025)]{seed15vl2025}
{ByteDance Seed}.
\newblock {Seed1.5-VL} technical report.
\newblock \emph{arXiv preprint arXiv:2505.07062}, 2025.

\bibitem[Chen et~al.(2024{\natexlab{a}})Chen, Dong, Shao, Hao, Yang, Su, and Zhu]{chen2024diffusion}
Huanran Chen, Yinpeng Dong, Shitong Shao, Zhongkai Hao, Xiao Yang, Hang Su, and Jun Zhu.
\newblock Diffusion models are certifiably robust classifiers.
\newblock In \emph{Advances in Neural Information Processing Systems}, 2024{\natexlab{a}}.

\bibitem[Chen et~al.(2023)Chen, Yu, Ge, Yao, Xie, Wu, Wang, Kwok, Luo, Lu, et~al.]{chen2023pixartalpha}
Junsong Chen, Jincheng Yu, Chongjian Ge, Lewei Yao, Enze Xie, Yue Wu, Zhongdao Wang, James Kwok, Ping Luo, Huchuan Lu, et~al.
\newblock {PixArt}-$\alpha$: Fast training of diffusion transformer for photorealistic text-to-image synthesis.
\newblock \emph{arXiv preprint arXiv:2310.00426}, 2023.

\bibitem[Chen et~al.(2024{\natexlab{b}})Chen, Wu, Wang, Su, Chen, Xing, Zhong, Zhang, Zhu, Lu, et~al.]{chen2024internvl}
Zhe Chen, Jiannan Wu, Wenhai Wang, Weijie Su, Guo Chen, Sen Xing, Muyan Zhong, Qinglong Zhang, Xizhou Zhu, Lewei Lu, et~al.
\newblock Internvl: Scaling up vision foundation models and aligning for generic visual-linguistic tasks.
\newblock In \emph{Proceedings of the IEEE/CVF Conference on Computer Vision and Pattern Recognition}, 2024{\natexlab{b}}.

\bibitem[Clark et~al.(2024)Clark, Vicol, Swersky, and Fleet]{clark2024direct}
Kevin Clark, Paul Vicol, Kevin Swersky, and David~J. Fleet.
\newblock Directly fine-tuning diffusion models on differentiable rewards.
\newblock In \emph{International Conference on Learning Representations}, 2024.

\bibitem[Domingo-Enrich et~al.(2025)Domingo-Enrich, Drozdzal, Karrer, and Chen]{domingoenrich2025adjoint}
Carles Domingo-Enrich, Michal Drozdzal, Brian Karrer, and Ricky T.~Q. Chen.
\newblock Adjoint matching: Fine-tuning flow and diffusion generative models with memoryless stochastic optimal control.
\newblock In \emph{International Conference on Learning Representations}, 2025.

\bibitem[Esser et~al.(2024)Esser, Kulal, Blattmann, Entezari, Muller, Saini, Levi, Lorenz, Sauer, Boesel, et~al.]{esser2024scaling}
Patrick Esser, Sumith Kulal, Andreas Blattmann, Rahim Entezari, Jonas Muller, Harry Saini, Yam Levi, Dominik Lorenz, Axel Sauer, Frederic Boesel, et~al.
\newblock Scaling rectified flow transformers for high-resolution image synthesis.
\newblock In \emph{International Conference on Machine Learning}, 2024.

\bibitem[Fan et~al.(2024)Fan, Watkins, Du, Liu, Ryu, Boutilier, Abbeel, Ghavamzadeh, Lee, and Lee]{fan2024dpok}
Ying Fan, Olivia Watkins, Yuqing Du, Hao Liu, Moonkyung Ryu, Craig Boutilier, Pieter Abbeel, Mohammad Ghavamzadeh, Kangwook Lee, and Kimin Lee.
\newblock {DPOK}: Reinforcement learning for fine-tuning text-to-image diffusion models.
\newblock \emph{arXiv preprint arXiv:2305.16381}, 2024.

\bibitem[Geng et~al.(2025)Geng, Deng, Bai, Kolter, and He]{geng2025meanflow}
Zhengyang Geng, Mingyang Deng, Xingjian Bai, J.~Zico Kolter, and Kaiming He.
\newblock Mean flows for one-step generative modeling.
\newblock \emph{arXiv preprint arXiv:2505.13447}, 2025.

\bibitem[Gholaminejad et~al.(2019)Gholaminejad, Keutzer, and Biros]{gholami2019anode}
Amir Gholaminejad, Kurt Keutzer, and George Biros.
\newblock {ANODE}: Unconditionally accurate memory-efficient gradients for neural {ODE}s.
\newblock In \emph{Proceedings of the Twenty-Eighth International Joint Conference on Artificial Intelligence}, pages 730--736, 2019.

\bibitem[Ghosh et~al.(2024)Ghosh, Hajishirzi, and Schmidt]{ghosh2024geneval}
Dhruba Ghosh, Hannaneh Hajishirzi, and Ludwig Schmidt.
\newblock {GenEval}: An object-focused framework for evaluating text-to-image alignment.
\newblock In \emph{Advances in Neural Information Processing Systems}, 2024.

\bibitem[Guo et~al.(2026)Guo, Wu, Liu, Gao, Ye, Yuan, Wang, Yu, and Huang]{guo2026editr1}
Hanzhong Guo, Jie Wu, Jie Liu, Yu~Gao, Zilyu Ye, Linxiao Yuan, Xionghui Wang, Yizhou Yu, and Weilin Huang.
\newblock Leveraging verifier-based reinforcement learning in image editing.
\newblock \emph{arXiv preprint arXiv:2604.27505}, 2026.

\bibitem[Hinton et~al.(2015)Hinton, Vinyals, and Dean]{hinton2015distilling}
Geoffrey Hinton, Oriol Vinyals, and Jeff Dean.
\newblock Distilling the knowledge in a neural network.
\newblock \emph{arXiv preprint arXiv:1503.02531}, 2015.

\bibitem[Ho et~al.(2022{\natexlab{a}})Ho, Chan, Saharia, Whang, Gao, Gritsenko, Kingma, Poole, Norouzi, Fleet, and Salimans]{ho2022imagenvideo}
Jonathan Ho, William Chan, Chitwan Saharia, Jay Whang, Ruiqi Gao, Alexey Gritsenko, Diederik~P. Kingma, Ben Poole, Mohammad Norouzi, David~J. Fleet, and Tim Salimans.
\newblock {Imagen Video}: High definition video generation with diffusion models.
\newblock \emph{arXiv preprint arXiv:2210.02303}, 2022{\natexlab{a}}.

\bibitem[Ho et~al.(2022{\natexlab{b}})Ho, Chan, Saharia, Whang, Gao, Gritsenko, Kingma, Poole, Norouzi, Fleet, and Salimans]{ho2022video}
Jonathan Ho, William Chan, Chitwan Saharia, Jay Whang, Ruiqi Gao, Alexey Gritsenko, Diederik~P. Kingma, Ben Poole, Mohammad Norouzi, David~J. Fleet, and Tim Salimans.
\newblock Video diffusion models.
\newblock In \emph{Advances in Neural Information Processing Systems}, 2022{\natexlab{b}}.

\bibitem[Huang et~al.(2024)Huang, He, Yu, Zhang, Si, Jiang, Zhang, Wu, Jin, Chanpaisit, et~al.]{huang2024vbench}
Ziqi Huang, Yinan He, Jiashuo Yu, Fan Zhang, Chenyang Si, Yuming Jiang, Yuanhan Zhang, Tianxing Wu, Qingyang Jin, Nattapol Chanpaisit, et~al.
\newblock {VBench}: Comprehensive benchmark suite for video generative models.
\newblock In \emph{Proceedings of the IEEE/CVF Conference on Computer Vision and Pattern Recognition}, pages 21807--21818, 2024.

\bibitem[Jiang et~al.(2025)Jiang, Liu, Wang, Wu, Li, Li, Jin, Liu, Li, Zhang, et~al.]{jiang2025dmdr}
Dengyang Jiang, Dongyang Liu, Zanyi Wang, Qilong Wu, Liuzhuozheng Li, Hengzhuang Li, Xin Jin, David Liu, Zhen Li, Bo~Zhang, et~al.
\newblock Distribution matching distillation meets reinforcement learning.
\newblock \emph{arXiv preprint arXiv:2511.13649}, 2025.

\bibitem[Kirstain et~al.(2023)Kirstain, Polyak, Singer, Matiana, Penna, and Levy]{kirstain2023pickapic}
Yuval Kirstain, Adam Polyak, Uriel Singer, Shahbuland Matiana, Joe Penna, and Omer Levy.
\newblock {Pick-a-Pic}: An open dataset of user preferences for text-to-image generation.
\newblock In \emph{Advances in Neural Information Processing Systems}, 2023.

\bibitem[Kong et~al.(2024)Kong, Tian, Zhang, Min, Dai, Zhou, Xiong, Li, Wu, Zhang, et~al.]{kong2024hunyuanvideo}
Weijie Kong, Qi~Tian, Zijian Zhang, Rox Min, Zuozhuo Dai, Jin Zhou, Jiangfeng Xiong, Xin Li, Bo~Wu, Jianwei Zhang, et~al.
\newblock {HunyuanVideo}: A systematic framework for large video generative models.
\newblock \emph{arXiv preprint arXiv:2412.03603}, 2024.

\bibitem[Li et~al.(2025)Li, Cui, Huang, Kong, Ma, Fan, Yang, Zhong, and Bo]{li2025mixgrpo}
Junzhe Li, Yutao Cui, Tao Huang, Weijie Kong, Yinping Ma, Chun Fan, Miles Yang, Zhao Zhong, and Liefeng Bo.
\newblock {MixGRPO}: Unlocking flow-based {GRPO} efficiency with mixed {ODE}-{SDE}.
\newblock \emph{arXiv preprint arXiv:2507.21802}, 2025.

\bibitem[Li et~al.(2026)Li, Zuo, He, Zhang, Xiao, Qian, Yu, Gao, Yang, Liu, et~al.]{rethinkingopd2026}
Yaxuan Li, Yuxin Zuo, Bingxiang He, Jinqian Zhang, Chaojun Xiao, Cheng Qian, Tianyu Yu, Huan-ang Gao, Wenkai Yang, Zhiyuan Liu, et~al.
\newblock Rethinking on-policy distillation of large language models: Phenomenology, mechanism, and recipe.
\newblock \emph{arXiv preprint arXiv:2604.13016}, 2026.

\bibitem[Liang et~al.(2026)Liang, Yang, Wu, Feng, and Zheng]{leapalign2026}
Zhanhao Liang, Tao Yang, Jie Wu, Chengjian Feng, and Liang Zheng.
\newblock {LeapAlign}: Post-training flow matching models at any generation step by building two-step trajectories.
\newblock \emph{arXiv preprint arXiv:2604.15311}, 2026.

\bibitem[Liu et~al.(2025)Liu, Liu, Liang, Li, Liu, Wang, Wan, Zhang, and Ouyang]{liu2025flowgrpo}
Jie Liu, Gongye Liu, Jiajun Liang, Yangguang Li, Jiaheng Liu, Xintao Wang, Pengfei Wan, Di~Zhang, and Wanli Ouyang.
\newblock {Flow-GRPO}: Training flow matching models via online {RL}.
\newblock \emph{arXiv preprint arXiv:2505.05470}, 2025.

\bibitem[Liu et~al.(2026)Liu, Liu, Liang, Yuan, Liu, Zheng, Wu, Wang, Xia, Wang, et~al.]{liu2025videoreward}
Jie Liu, Gongye Liu, Jiajun Liang, Ziyang Yuan, Xiaokun Liu, Mingwu Zheng, Xiele Wu, Qiulin Wang, Menghan Xia, Xintao Wang, et~al.
\newblock Improving video generation with human feedback.
\newblock In \emph{Advances in Neural Information Processing Systems}, volume~38, pages 82155--82192, 2026.

\bibitem[Liu et~al.(2022)Liu, Gong, and Liu]{liu2022flow}
Xingchao Liu, Chengyue Gong, and Qiang Liu.
\newblock Flow straight and fast: Learning to generate and transfer data with rectified flow.
\newblock \emph{arXiv preprint arXiv:2209.03003}, 2022.

\bibitem[Luo et~al.(2023)Luo, Tan, Huang, Li, and Zhao]{luo2023latent}
Simian Luo, Yiqin Tan, Longbo Huang, Jian Li, and Hang Zhao.
\newblock Latent consistency models: Synthesizing high-resolution images with few-step inference.
\newblock \emph{arXiv preprint arXiv:2310.04378}, 2023.

\bibitem[Luo(2024)]{luo2024diffinstructpp}
Weijian Luo.
\newblock {Diff-Instruct++}: Training one-step text-to-image generator model to align with human preferences.
\newblock \emph{arXiv preprint arXiv:2410.18881}, 2024.

\bibitem[Ma et~al.(2025)Ma, Wu, Sun, and Li]{ma2025hpsv3}
Yuhang Ma, Xiaoshi Wu, Keqiang Sun, and Hongsheng Li.
\newblock {HPSv3}: Towards wide-spectrum human preference score.
\newblock In \emph{2025 IEEE/CVF International Conference on Computer Vision (ICCV)}, pages 15086--15095. IEEE, 2025.

\bibitem[Miao et~al.(2024)Miao, Yang, Lin, Wang, Liu, Wang, and Qiu]{miao2024pso}
Zichen Miao, Zhengyuan Yang, Kevin Lin, Ze~Wang, Zicheng Liu, Lijuan Wang, and Qiang Qiu.
\newblock Tuning timestep-distilled diffusion model using pairwise sample optimization.
\newblock \emph{arXiv preprint arXiv:2410.03190}, 2024.

\bibitem[Onken and Ruthotto(2020)]{onken2020discretize}
Derek Onken and Lars Ruthotto.
\newblock Discretize-optimize vs. optimize-discretize for time-series regression and continuous normalizing flows.
\newblock \emph{arXiv preprint arXiv:2005.13420}, 2020.

\bibitem[Ouyang et~al.(2022)Ouyang, Wu, Jiang, Almeida, Wainwright, Mishkin, Zhang, Agarwal, Slama, Ray, et~al.]{ouyang2022training}
Long Ouyang, Jeffrey Wu, Xu~Jiang, Diogo Almeida, Carroll~L. Wainwright, Pamela Mishkin, Chong Zhang, Sandhini Agarwal, Katarina Slama, Alex Ray, et~al.
\newblock Training language models to follow instructions with human feedback.
\newblock In \emph{Advances in Neural Information Processing Systems}, 2022.

\bibitem[Peebles and Xie(2023)]{peebles2023scalable}
William Peebles and Saining Xie.
\newblock Scalable diffusion models with transformers.
\newblock In \emph{Proceedings of the IEEE/CVF International Conference on Computer Vision}, 2023.

\bibitem[Podell et~al.(2023)Podell, English, Lacey, Blattmann, Dockhorn, Muller, Penna, and Rombach]{podell2023sdxl}
Dustin Podell, Zion English, Kyle Lacey, Andreas Blattmann, Tim Dockhorn, Jonas Muller, Joe Penna, and Robin Rombach.
\newblock {SDXL}: Improving latent diffusion models for high-resolution image synthesis.
\newblock \emph{arXiv preprint arXiv:2307.01952}, 2023.

\bibitem[Prabhudesai et~al.(2023)Prabhudesai, Goyal, Pathak, and Fragkiadaki]{prabhudesai2023alignprop}
Mihir Prabhudesai, Anirudh Goyal, Deepak Pathak, and Katerina Fragkiadaki.
\newblock Aligning text-to-image diffusion models with reward backpropagation.
\newblock \emph{arXiv preprint arXiv:2310.03739}, 2023.

\bibitem[Rafailov et~al.(2023)Rafailov, Sharma, Mitchell, Manning, Ermon, and Finn]{rafailov2023direct}
Rafael Rafailov, Archit Sharma, Eric Mitchell, Christopher~D. Manning, Stefano Ermon, and Chelsea Finn.
\newblock Direct preference optimization: Your language model is secretly a reward model.
\newblock In \emph{Advances in Neural Information Processing Systems}, 2023.

\bibitem[Ramesh et~al.(2022)Ramesh, Dhariwal, Nichol, Chu, and Chen]{ramesh2022hierarchical}
Aditya Ramesh, Prafulla Dhariwal, Alex Nichol, Casey Chu, and Mark Chen.
\newblock Hierarchical text-conditional image generation with {CLIP} latents.
\newblock \emph{arXiv preprint arXiv:2204.06125}, 2022.

\bibitem[Ren et~al.(2025)Ren, Lidard, Ankile, Simeonov, Agrawal, Majumdar, Burchfiel, Dai, and Simchowitz]{ren2025dppo}
Allen~Z. Ren, Justin Lidard, Lars Ankile, Anthony Simeonov, Pulkit Agrawal, Anirudha Majumdar, Benjamin Burchfiel, Hongkai Dai, and Max Simchowitz.
\newblock Diffusion policy policy optimization.
\newblock In \emph{International Conference on Learning Representations}, 2025.

\bibitem[Ren et~al.(2024)Ren, Xia, Lu, Zhang, Wu, Xie, Wang, and Xiao]{ren2024hypersd}
Yuxi Ren, Xin Xia, Yanzuo Lu, Jiacheng Zhang, Jie Wu, Pan Xie, Xing Wang, and Xuefeng Xiao.
\newblock {Hyper-SD}: Trajectory segmented consistency model for efficient image synthesis.
\newblock In \emph{Advances in Neural Information Processing Systems}, 2024.

\bibitem[Rombach et~al.(2022)Rombach, Blattmann, Lorenz, Esser, and Ommer]{rombach2022latent}
Robin Rombach, Andreas Blattmann, Dominik Lorenz, Patrick Esser, and Bjorn Ommer.
\newblock High-resolution image synthesis with latent diffusion models.
\newblock In \emph{Proceedings of the IEEE/CVF Conference on Computer Vision and Pattern Recognition}, 2022.

\bibitem[Rusu et~al.(2015)Rusu, Colmenarejo, Gulcehre, Desjardins, Kirkpatrick, Pascanu, Mnih, Kavukcuoglu, and Hadsell]{rusu2015policy}
Andrei~A. Rusu, Sergio~Gomez Colmenarejo, Caglar Gulcehre, Guillaume Desjardins, James Kirkpatrick, Razvan Pascanu, Volodymyr Mnih, Koray Kavukcuoglu, and Raia Hadsell.
\newblock Policy distillation.
\newblock \emph{arXiv preprint arXiv:1511.06295}, 2015.

\bibitem[Saharia et~al.(2022)Saharia, Chan, Saxena, Li, Whang, Denton, Ghasemipour, Gontijo~Lopes, Karagol~Ayan, Salimans, et~al.]{saharia2022photorealistic}
Chitwan Saharia, William Chan, Saurabh Saxena, Lala Li, Jay Whang, Emily Denton, Kamyar Ghasemipour, Raphael Gontijo~Lopes, Burcu Karagol~Ayan, Tim Salimans, et~al.
\newblock Photorealistic text-to-image diffusion models with deep language understanding.
\newblock In \emph{Advances in Neural Information Processing Systems}, 2022.

\bibitem[Salimans and Ho(2022)]{salimans2022progressive}
Tim Salimans and Jonathan Ho.
\newblock Progressive distillation for fast sampling of diffusion models.
\newblock In \emph{International Conference on Learning Representations}, 2022.

\bibitem[Sauer et~al.(2024)Sauer, Lorenz, Blattmann, and Rombach]{sauer2024adversarial}
Axel Sauer, Dominik Lorenz, Andreas Blattmann, and Robin Rombach.
\newblock Adversarial diffusion distillation.
\newblock In \emph{European Conference on Computer Vision}, 2024.

\bibitem[Schulman et~al.(2017)Schulman, Wolski, Dhariwal, Radford, and Klimov]{schulman2017ppo}
John Schulman, Filip Wolski, Prafulla Dhariwal, Alec Radford, and Oleg Klimov.
\newblock Proximal policy optimization algorithms.
\newblock \emph{arXiv preprint arXiv:1707.06347}, 2017.

\bibitem[{Seed Vision Team}(2025{\natexlab{a}})]{seedance2025}
{Seed Vision Team}.
\newblock Seedance 1.0: Exploring the boundaries of video generation models.
\newblock \emph{arXiv preprint arXiv:2506.09113}, 2025{\natexlab{a}}.

\bibitem[{Seed Vision Team}(2025{\natexlab{b}})]{seedream2025}
{Seed Vision Team}.
\newblock Seedream 3.0 technical report.
\newblock \emph{arXiv preprint arXiv:2504.11346}, 2025{\natexlab{b}}.

\bibitem[Shao et~al.(2024)Shao, Wang, Zhu, Xu, Song, Bi, Zhang, Zhang, Li, Wu, et~al.]{shao2024grpo}
Zhihong Shao, Peiyi Wang, Qihao Zhu, Runxin Xu, Junxiao Song, Xiao Bi, Haowei Zhang, Mingchuan Zhang, YK~Li, Yang Wu, et~al.
\newblock {DeepSeekMath}: Pushing the limits of mathematical reasoning in open language models.
\newblock \emph{arXiv preprint arXiv:2402.03300}, 2024.

\bibitem[Singer et~al.(2022)Singer, Polyak, Hayes, Yin, An, Zhang, Hu, Yang, Ashual, Gafni, et~al.]{singer2022makeavideo}
Uriel Singer, Adam Polyak, Thomas Hayes, Xiaoyue Yin, Jie An, Songyang Zhang, Qiyuan Hu, Harry Yang, Oron Ashual, Oran Gafni, et~al.
\newblock {Make-A-Video}: Text-to-video generation without text-video data.
\newblock \emph{arXiv preprint arXiv:2209.14792}, 2022.

\bibitem[Song and Zheng(2026)]{opdsurvey2026}
Mingyang Song and Mao Zheng.
\newblock A survey of on-policy distillation for large language models.
\newblock \emph{arXiv preprint arXiv:2604.00626}, 2026.

\bibitem[Song et~al.(2023)Song, Dhariwal, Chen, and Sutskever]{song2023consistency}
Yang Song, Prafulla Dhariwal, Mark Chen, and Ilya Sutskever.
\newblock Consistency models.
\newblock In \emph{International Conference on Machine Learning}, 2023.

\bibitem[Sutton and Barto(2018)]{sutton2018reinforcement}
Richard~S. Sutton and Andrew~G. Barto.
\newblock \emph{Reinforcement Learning: An Introduction}.
\newblock MIT Press, second edition, 2018.

\bibitem[{Team Seedance}(2026)]{seedance2026seedance}
{Team Seedance}.
\newblock Seedance 2.0: Advancing video generation for world complexity.
\newblock \emph{arXiv preprint arXiv:2604.14148}, 2026.

\bibitem[Wallace et~al.(2024)Wallace, Dang, Rafailov, Zhou, Lou, Purushwalkam, Ermon, Xiong, Joty, and Naik]{wallace2024diffusiondpo}
Bram Wallace, Meihua Dang, Rafael Rafailov, Linqi Zhou, Aaron Lou, Senthil Purushwalkam, Stefano Ermon, Caiming Xiong, Shafiq Joty, and Nikhil Naik.
\newblock Diffusion model alignment using direct preference optimization.
\newblock In \emph{Proceedings of the IEEE/CVF Conference on Computer Vision and Pattern Recognition}, 2024.

\bibitem[{Wan Team} et~al.(2025){Wan Team}, Wang, Ai, Wen, Mao, Xie, Chen, Yu, Zhao, Yang, et~al.]{wan2025wan}
{Wan Team}, Ang Wang, Baole Ai, Bin Wen, Chaojie Mao, Chen-Wei Xie, Di~Chen, Feiwu Yu, Haiming Zhao, Jianxiao Yang, et~al.
\newblock {Wan}: Open and advanced large-scale video generative models.
\newblock \emph{arXiv preprint arXiv:2503.20314}, 2025.

\bibitem[Wang et~al.(2025)Wang, Zang, Li, Jin, and Wang]{unifiedreward2025}
Yibin Wang, Yuhang Zang, Hao Li, Cheng Jin, and Jiaqi Wang.
\newblock Unified reward model for multimodal understanding and generation.
\newblock \emph{arXiv preprint arXiv:2503.05236}, 2025.

\bibitem[Wu et~al.(2025{\natexlab{a}})Wu, Li, Zhou, Lin, Gao, Yan, Yin, Bai, Xu, Chen, et~al.]{wu2025qwen}
Chenfei Wu, Jiahao Li, Jingren Zhou, Junyang Lin, Kaiyuan Gao, Kun Yan, Sheng-ming Yin, Shuai Bai, Xiao Xu, Yilei Chen, et~al.
\newblock Qwen-image technical report.
\newblock \emph{arXiv preprint arXiv:2508.02324}, 2025{\natexlab{a}}.

\bibitem[Wu et~al.(2025{\natexlab{b}})Wu, Gao, Ye, Li, Li, Guo, Liu, Xue, Hou, Liu, et~al.]{rewarddance2025}
Jie Wu, Yu~Gao, Zilyu Ye, Ming Li, Liang Li, Hanzhong Guo, Jie Liu, Zeyue Xue, Xiaoxia Hou, Wei Liu, et~al.
\newblock Rewarddance: Reward scaling in visual generation.
\newblock \emph{arXiv preprint arXiv:2509.08826}, 2025{\natexlab{b}}.

\bibitem[Wu et~al.(2023)Wu, Hao, Sun, Chen, Zhu, Zhao, and Li]{wu2023hps}
Xiaoshi Wu, Yixiao Hao, Keqiang Sun, Yixiong Chen, Feng Zhu, Rui Zhao, and Hongsheng Li.
\newblock Human preference score v2: A solid benchmark for evaluating human preferences of text-to-image synthesis.
\newblock \emph{arXiv preprint arXiv:2306.09341}, 2023.

\bibitem[Wu et~al.(2024)Wu, Hao, Zhang, Sun, Huang, Song, Liu, and Li]{wu2024drtune}
Xiaoshi Wu, Yiming Hao, Manyuan Zhang, Keqiang Sun, Zhaoyang Huang, Guanglu Song, Yu~Liu, and Hongsheng Li.
\newblock Deep reward supervisions for tuning text-to-image diffusion models.
\newblock In \emph{European Conference on Computer Vision}, 2024.

\bibitem[Xu et~al.(2023)Xu, Liu, Wu, Tong, Li, Ding, Tang, and Dong]{xu2023imagereward}
Jiazheng Xu, Xiao Liu, Yuchen Wu, Yuxuan Tong, Qinkai Li, Ming Ding, Jie Tang, and Yuxiao Dong.
\newblock {ImageReward}: Learning and evaluating human preferences for text-to-image generation.
\newblock In \emph{Advances in Neural Information Processing Systems}, 2023.

\bibitem[Xue et~al.(2025)Xue, Wu, Gao, Kong, Zhu, Chen, Liu, Liu, Guo, Huang, et~al.]{dancegrpo2025}
Zeyue Xue, Jie Wu, Yu~Gao, Fangyuan Kong, Lingting Zhu, Mengzhao Chen, Zhiheng Liu, Wei Liu, Qiushan Guo, Weilin Huang, et~al.
\newblock {DanceGRPO}: Unleashing {GRPO} on visual generation.
\newblock \emph{arXiv preprint arXiv:2505.07818}, 2025.

\bibitem[Yang et~al.(2024{\natexlab{a}})Yang, Tao, Lyu, Ge, Chen, Li, Shen, Zhu, and Li]{yang2024d3po}
Kai Yang, Jian Tao, Jiafei Lyu, Chunjiang Ge, Jiaxin Chen, Qimai Li, Weihan Shen, Xiaolong Zhu, and Xiu Li.
\newblock Using human feedback to fine-tune diffusion models without any reward model.
\newblock In \emph{Proceedings of the IEEE/CVF Conference on Computer Vision and Pattern Recognition}, 2024{\natexlab{a}}.

\bibitem[Yang et~al.(2024{\natexlab{b}})Yang, Teng, Zheng, Ding, Huang, Xu, Yang, Hong, Zhang, Feng, et~al.]{yang2024cogvideox}
Zhuoyi Yang, Jiayan Teng, Wendi Zheng, Ming Ding, Shiyu Huang, Jiazheng Xu, Yuanming Yang, Wenyi Hong, Xiaohan Zhang, Guanyu Feng, et~al.
\newblock {CogVideoX}: Text-to-video diffusion models with an expert transformer.
\newblock \emph{arXiv preprint arXiv:2408.06072}, 2024{\natexlab{b}}.

\bibitem[Yin et~al.(2024)Yin, Gharbi, Park, Zhang, Shechtman, Durand, and Freeman]{yin2024improved}
Tianwei Yin, Michael Gharbi, Taesung Park, Richard Zhang, Eli Shechtman, Fredo Durand, and William~T. Freeman.
\newblock Improved distribution matching distillation for fast image synthesis.
\newblock In \emph{Advances in Neural Information Processing Systems}, 2024.

\bibitem[Zheng et~al.(2026)Zheng, Chen, Ye, Wang, Zhang, Jiang, Su, Ermon, Zhu, and Liu]{zheng2025diffusionnft}
Kaiwen Zheng, Huayu Chen, Haotian Ye, Haoxiang Wang, Qinsheng Zhang, Kai Jiang, Hang Su, Stefano Ermon, Jun Zhu, and Ming-Yu Liu.
\newblock {DiffusionNFT}: Online diffusion reinforcement with forward process.
\newblock In \emph{International Conference on Learning Representations}, 2026.

\end{thebibliography}
